\documentclass{article}

\usepackage[eandd, preprint]{neurips_2026}

\usepackage[utf8]{inputenc}
\usepackage[T1]{fontenc}
\usepackage{hyperref}
\usepackage{url}
\usepackage{booktabs}
\usepackage{amsmath,amssymb,amsfonts,amsthm,mathtools}
\usepackage{microtype}
\usepackage{xcolor}
\usepackage{multirow}
\usepackage[inline]{enumitem}
\usepackage{subcaption}
\usepackage{graphicx}
\usepackage{pifont}
\usepackage{xcolor}
\usepackage[most]{tcolorbox}
\definecolor{boxblue}{RGB}{235, 245, 255} 

\newcommand{\benchname}{\textsc{LowRankArena}}

\usepackage{makecell}

\usepackage{array}
\newcolumntype{L}[1]{>{\raggedright\arraybackslash}p{#1}}

\usepackage{booktabs}
\usepackage{longtable}
\usepackage{array}
\usepackage{ragged2e}
\usepackage{makecell}
\usepackage{pdflscape}
\usepackage{tabularx}
\usepackage{threeparttable}
\usepackage{wrapfig}

\usepackage{booktabs}
\usepackage[table]{xcolor}
\usepackage{multirow}
\usepackage{siunitx}

\definecolor{LRHeader}{RGB}{232,241,250}
\definecolor{LRDense}{RGB}{247,247,247}

\usepackage{pifont}

\title{\benchname: A Standardized Evaluation Platform for SVD-Based LLM Compression}

\author{%
  \textbf{Zishan Shao}$^{1,}$\thanks{Equal contribution.}, 
  \textbf{Lixun Zhang}$^{1,*}$, 
  \textbf{Kangning Cui}$^{2,*}$, 
  \textbf{Wenhao Wu}$^{1,*}$, 
  \textbf{Jinhee Kim}$^1$, 
  \textbf{Yixiao Wang}$^1$, \\
  \textbf{Ting Jiang}$^1$, 
  \textbf{Hancheng Ye}$^1$, 
  \textbf{Qinsi Wang}$^1$, 
  \textbf{Fan Yang}$^2$, 
  \textbf{Danyang Zhuo}$^1$, 
  \textbf{Yiran Chen}$^1$, 
  \textbf{Hai Li}$^1$ \\[0.5em]
  $^1$Duke University \quad $^2$Wake Forest University \\[0.3em]
  \small Correspondence to: \texttt{zishan.shao@duke.edu}
}

\begin{document}

\maketitle

\begin{abstract}

SVD-based low-rank compression has become a fast-growing direction for reducing the memory and computational cost of large language models (LLMs). However,  meaningful comparison across existing studies remains difficult as prior evaluations use varied benchmarks, inconsistent ratios, and diverse setups, often failing to isolate low-rank effects from auxiliary techniques. As a result, it remains unclear whether reported gains reflect method-level improvements or differences in evaluation protocol. This lack of comparability highlights the need for a unified, reproducible evaluation platform. To address this problem, we present \textbf{\textsc{LowRankArena}}, a standardized evaluation platform for SVD-based LLM compression. \textsc{LowRankArena} unifies task versions, uniform-precision compression budgets, comparison regimes, and inference measurements, and provides a reproducible pipeline with over 3 TiB released compressed checkpoints. Using \textsc{LowRankArena}, our aligned audit of five representative SVD methods reveals that prior findings are highly conditional under standardized protocols: clear leaders and performance tiers shift across backbones and keep ratios, multiple-choice accuracy can hide large perplexity degradation, and nominal low-rank savings yield workload-dependent and often limited end-to-end speedups. Our code is available at: \href{https://github.com/Zishan-Shao/lowrankarena.git}{https://github.com/Zishan-Shao/lowrankarena.git}.

\end{abstract}



\section{Introduction}
\label{sec:intro}

SVD-based low-rank compression has attracted growing interest as a simple approach for reducing the memory and computational costs of large language models (LLMs). By factorizing dense weight matrices into low-rank components, it provides a broadly applicable form of checkpoint compression, and recent work has reported encouraging results in both accuracy retention and inference efficiency.

However, these results remain difficult to compare fairly. Existing papers~\cite{abbasi2026zssvd,gao2026dfsvd,hu2026saes,sinha2026aasvd,li2025adasvd,hsu2022fwsvd,ding2025dipsvd,wang2025svdllm,wang2025svdllmv2,lin2024modegpt,wang2025dobi} often differ in benchmark suites, task versions, model backbones, and inference backend. Even the definition of ``compression ratio'' remains unstandardized; in several instances, gains from mixed-precision quantization or remapping~\cite{wang2025dobi,abbasi2026zssvd,sinha2026aasvd} are bundled into the low-rank budget, which obscures the true effectiveness of the SVD-based decomposition and complicates comparisons with established baselines like structured pruning. Efficiency reporting is equally fragmented, with many works prioritizing prefill-centric execution profiles or isolated layer-level latency measurements over end-to-end inference throughput in practical deployment scenarios.

Reproducibility further compounds this problem. Re-running SVD compression baselines across model families, keep ratios, task suites, and inference configurations requires substantial GPU resources. As a result, researchers often rely on reported figures rather than direct re-evaluation under a shared pipeline. This dependence makes it difficult to determine whether observed gains come from the compression algorithm itself or from experimental setup differences. Consequently, the community lacks a unified evaluation platform and reproducible checkpoints for answering a central question:

\begin{tcolorbox}[
    enhanced,
    frame hidden,
    borderline west={2pt}{0pt}{black!70},
    colback=gray!10,
    boxrule=0pt,
    sharp corners,
    left=8pt,
    top=6pt,
    bottom=6pt,
    before skip=12pt,
    after skip=12pt
]
\textbf{\textit{When SVD-based LLM compression methods are evaluated under a fixed uniform-precision parameter budget, aligned task versions, and a shared inference stack, which conclusions about recent progress still hold?}}
\end{tcolorbox}

To answer this question, we present the \textbf{\benchname}, a unified evaluation platform for SVD-based LLM compression. \benchname \, standardizes task versions, model settings, compression budgets, and end-to-end inference measurements. To support reproducible future comparisons, we release more than 3~TiB of reproducible artifacts, including the full evaluation pipeline and an extensive collection of compressed checkpoints.

Using \benchname, we revisit existing SVD methods under aligned settings and find that part of the apparent progress in isolated evaluations reflects protocol sensitivity. Our results show \textbf{\emph{no stable, architecture-invariant ordering once evaluation axes are aligned, with relative strengths depending heavily on the backbone, keep ratio, and metric group.}} Regarding efficiency, SVD benefits are highly workload-dependent: gains in compute-bound scenarios like large-batch prefill diminish in generation-heavy tasks where decoding remains a bottleneck. This suggests that \textbf{current speedups are concentrated in \emph{compute-bound stages}} rather than the entire generation lifecycle.

\begin{figure}
    \centering
    \includegraphics[width=0.99\linewidth]{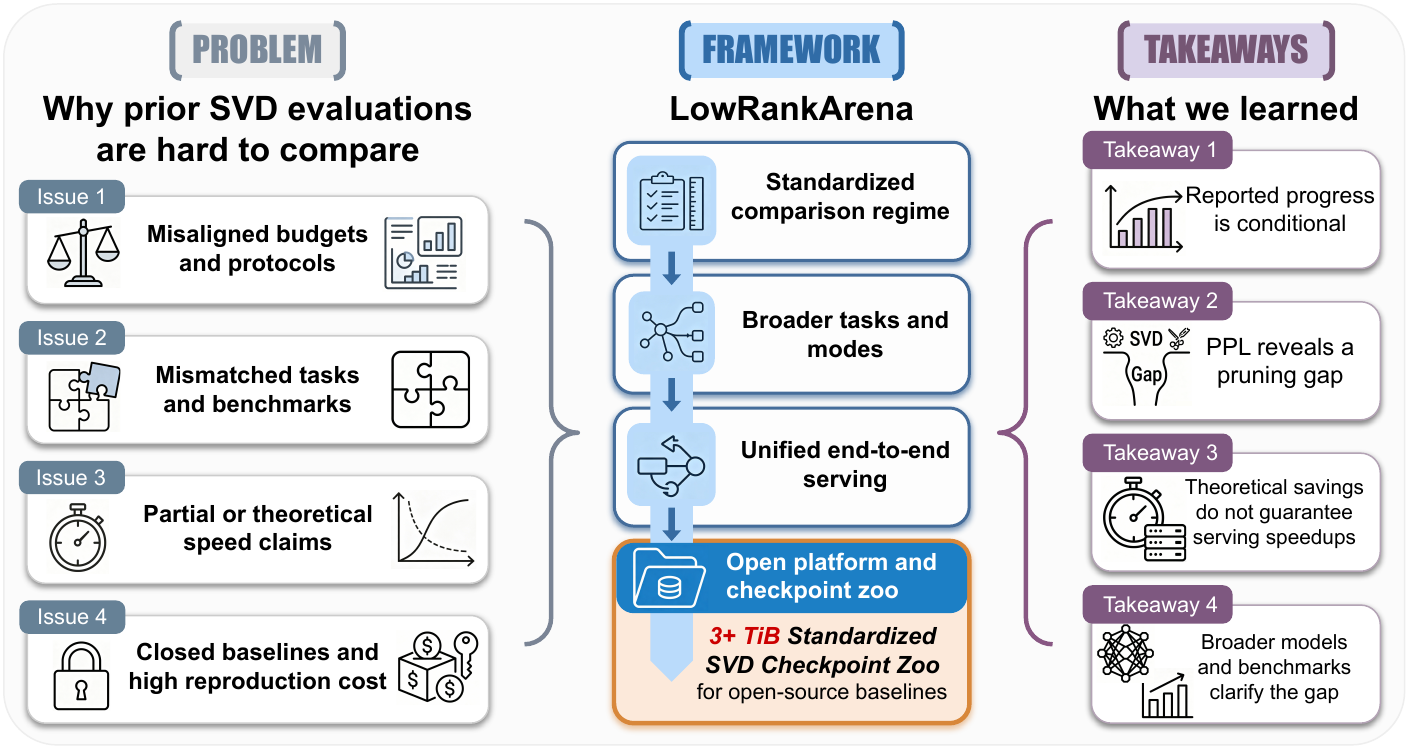}
    \caption{\textbf{Overview of \benchname.} \textbf{Left:} The current evaluation of SVD-based LLM compression suffers from inconsistent settings and high reproduction costs. \textbf{Middle:} \benchname \, addresses this by providing a standardized, fully reproducible evaluation platform with over 3~TiB of released checkpoints. \textbf{Right:} By auditing representative publicly reproducible methods under this unified protocol, we characterize the reproducible SVD-compression frontier, highlighting unstable rankings, no consistent advantage over structured pruning, and a mismatch between theoretical and practical end-to-end speedups. \textbf{Checkpoints:} \url{https://huggingface.co/Duke-CEI-SVD/LowRankArena}. 
    }
    \label{fig:pipeline}
\end{figure}



\section{The \textsc{LowRankArena} Platform}
\label{sec:lowrankarena}



We now describe \textsc{LowRankArena}, our unified evaluation platform for SVD-based low-rank LLM compression.  \textsc{LowRankArena} is designed not as a collection of isolated benchmark results, but as a comparison protocol for \textbf{\textit{turning heterogeneous SVD claims into matched and reproducible evaluations.}} It fixes the evaluation assumptions that most often vary across prior work, separates pure low-rank compression from auxiliary techniques such as mixed precision, weight remapping, or runtime-specific execution policies, and provides released artifacts that serve as common reference points for future methods. These design choices make the results in the rest of the paper easier to interpret as method-level differences rather than artifacts of incompatible evaluation protocols.

\begin{figure}[t]
    \centering
    \includegraphics[width=0.99\linewidth]{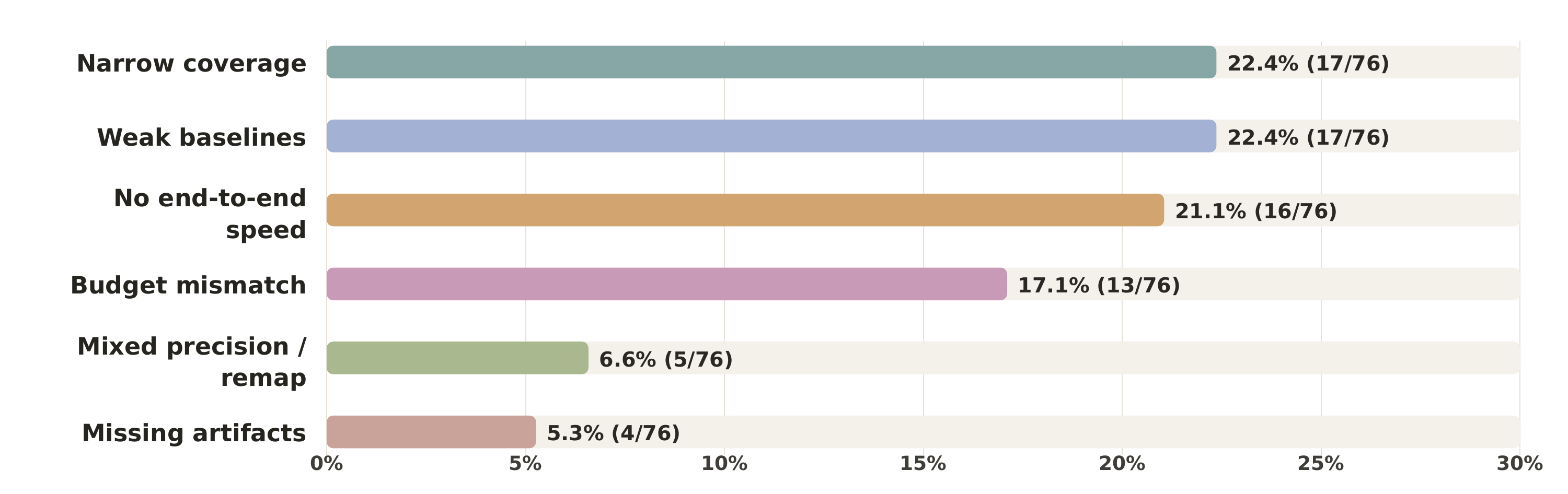}
    \caption{\textbf{Recurring reviewer concerns in recent SVD evaluations.}
    Bars show the share of public OpenReview reviewer/Area Chair notes ($n=76$) mentioning each concern category, highlighting systematic evaluation fragmentation and motivating LowRankArena.}
    \label{fig:reviewer_concerns}
\end{figure}

\subsection{Motivation and Design}
\label{sec:motivation}
\paragraph{Why current SVD evaluations are hard to compare.}

Our benchmark design is guided by a systematic audit of 76 public reviewer notes on recent SVD-based LLM compression papers (Fig.~\ref{fig:reviewer_concerns}). The audit shows that evaluation fragmentation is not merely anecdotal, but a \textbf{\emph{recurring obstacle to assessing field-level progress}}. The most common concerns can be grouped into several recurring themes. First, \textit{narrow coverage} and \textit{budget mismatches} arise because recent papers differ in model families, task versions, and budget definitions~\cite{wang2025svdllm, wang2025dobi, yuan2023asvd}. Second, the frequent lack of \textit{end-to-end speed} reporting~\cite{lin2024modegpt} limits practical assessment of deployment efficiency. Third, \textit{mixed-precision} and remapping variants (quantization-assisted mixed-precision encoding tricks)~\cite{wang2025dobi,sinha2026aasvd,abbasi2026zssvd} are often reported under the same labels as pure low-rank compression~\cite{abbasi2026zssvd}, despite relying on different hardware and runtime assumptions. Finally, a significant \textit{transparency gap} persists: missing artifacts and weak baseline coverage in recent research (2025--2026) make later comparisons depend on \emph{non-verifiable, author-reported results} rather than matched re-evaluation~\cite{hu2026saes,gao2026dfsvd,gao2025gfsvd,wang2025basis,wang2025dobi}.


\paragraph{What \textsc{LowRankArena} standardizes.}

\textsc{LowRankArena} addresses these comparability issues through a systematic evaluation framework that standardizes evaluation benchmarks, compression budgets, and inference environments. Instead of aggregating numbers from incompatible settings, it evaluates each method through the same artifact interface, task definitions, budget axis, and inference stack. This design keeps model quality, compression budget, and deployment efficiency as separately controlled dimensions, so that changes in performance can be traced to the compression method rather than to hidden differences in evaluation setup. To support reproducible comparison, we release standardized method adapters and over 3~TiB of compressed checkpoints. These evaluation pipeline features provide common reference points for evaluating future methods against shared baselines and verifiable model states.

\paragraph{Comparison regimes.}

\textsc{LowRankArena} deliberately separates headline comparisons from auxiliary audits. Because uniform-precision, mixed-precision, remapping-based, and runtime-adaptive methods rely on different deployment assumptions, placing them on a single leaderboard would confound algorithmic low-rank quality with implementation-specific advantages. We therefore define uniform-precision SVD as the primary regime, evaluated under the shared protocol and compared with structured pruning on the same budget axis and precision. All main leaderboard results and central claims are based on this regime. Mixed-precision, remapping, and runtime-adaptive variants are not merged into the primary leaderboard. Instead, we report auxiliary audits on artifact coverage, large-model compression feasibility, and cross-device inference robustness to clarify which conclusions are method-level and which depend on implementation or deployment assumptions.

\begin{figure}[ht]
    \centering
    \includegraphics[width=0.99\linewidth]{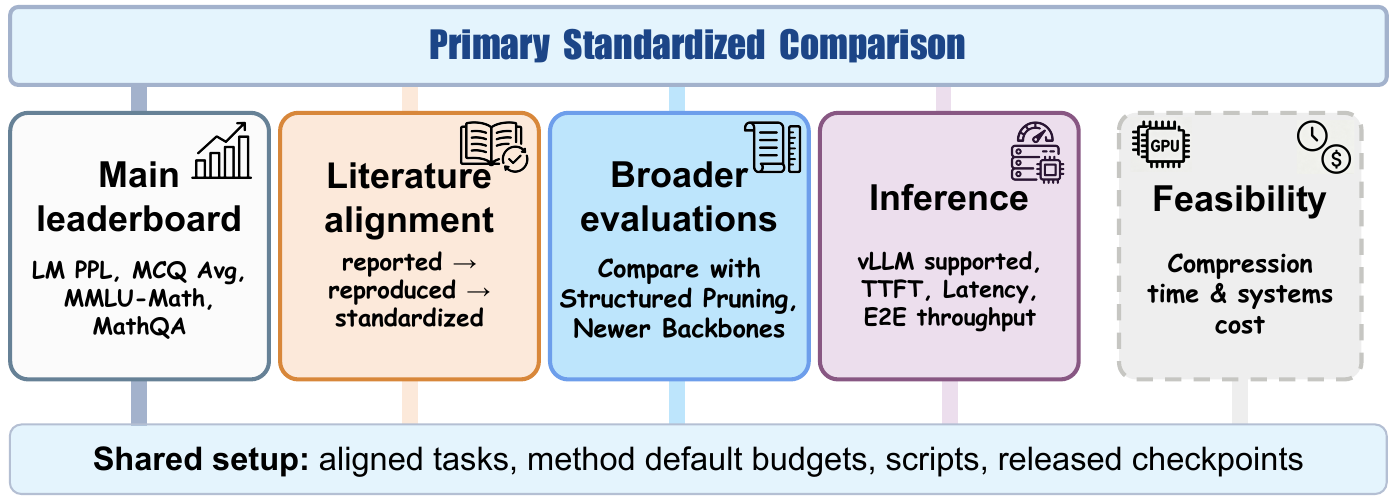}
    \caption{\textbf{Evaluation scope of LowRankArena.} Our platform standardizes SVD assessment across five tracks: (1) \textbf{Main leaderboard} for reasoning/PPL; (2) \textbf{Literature alignment} for prior claims; (3) \textbf{Broader evaluations} for architecture and baseline scaling; (4) \textbf{inference} for end-to-end speed; and (5) \textbf{Feasibility} for compression costs. All tracks utilize a unified setup with matched budgets.}
 \label{fig:placeholder}
\end{figure}

\subsection{Models, Tasks, and Inference}

\benchname{} standardizes low-rank compression along four controlled axes:
model coverage, task coverage, inference measurement, and compression budget.
This keeps capability retention, compression level, and deployable efficiency
separable, rather than folding them into a single opaque score.


\begin{tcolorbox}[
    sharp corners,
    boxrule=0.4pt,
    colback=gray!3,
    colframe=black!35,
    left=6pt,
    right=6pt,
    top=5pt,
    bottom=5pt
]
\textbf{Standardized Keep Ratio.}
For compressed linear layers $\mathcal{L}$,
\[
r =
\frac{\sum_{l \in \mathcal{L}} k_l (m_l + n_l)}
     {\sum_{l \in \mathcal{L}} m_l n_l},
\qquad
\mathrm{prec}(\hat{W}_l)=\mathrm{prec}(W_l).
\]
Here $W_l \in \mathbb{R}^{m_l \times n_l}$ and $k_l$ is the retained rank and $\operatorname{prec}(\cdot)$ denotes the bit-width.
Thus, fixed-$r$ comparisons isolate pure subspace selection rather than
bit-width reduction, or dense fallback.
\end{tcolorbox}

\paragraph{Models and baselines.}
We evaluate legacy LLaMA backbones
(LLaMA-1/2-7B~\cite{touvron2023llama,touvron2023llama2}) together with newer
compression targets, including Llama-3.1~\cite{grattafiori2024llama3} and
Qwen3~\cite{yang2025qwen3}.  The standardized suite includes publicly
reproducible SVD-style methods: ASVD~\cite{yuan2023asvd},
SVD-LLM~\cite{wang2025svdllm}, DoBi-SVD~\cite{wang2025dobi}, Basis
Sharing~\cite{wang2025basis}, and MoDeGPT~\cite{lin2024modegpt}.  Unless
noted, methods are evaluated at $r\in\{0.8,0.6,0.4\}$ with aligned model, task, keep-ratio, precision, and evaluation settings, while retaining method default calibration. All primary comparisons exclude separately added post-compression recovery. 

\paragraph{Tasks and metrics.}
We pair answer-selection benchmarks with generative language-modeling metrics. Using LM-Eval-Harness v0.4.11~\cite{biderman2024lm_eval,eleutherai2026lm_eval_v0411}, we report zero-shot accuracy over seven standard multiple-choice tasks, and separately measure perplexity on WikiText-2~\cite{merity2016pointer} and C4~\cite{raffel2020exploring}.  For base models, we also evaluate mathematical and disciplinary reasoning with MathQA~\cite{amini2019mathqa} and MMLU-Math~\cite{hendrycks2021mmlu} (5-shot); for instruction-tuned models, we include MMLU-Pro~\cite{wang2024mmlupro} (5-shot), GSM8K~\cite{cobbe2021gsm8k} (8-shot), and IFEval~\cite{zhou2023ifeval}.  Full task aliases, prompts, shot counts, and scoring rules are provided in Appendix~\ref{sec:appendix_tasks}. For backend we uses the vLLM~\cite{kwon2023efficient} with verison 0.18.1. 

\paragraph{Inference measurements.}
We measure deployable efficiency with exported checkpoints served through a single vLLM path~\cite{kwon2023efficient}.  Within each table, hardware, precision, tensor parallelism, scheduler settings, prompt/output lengths, and request arrival process are fixed across methods.  We report matched time-to-first-token (TTFT), inter-token latency, end-to-end latency (E2E), and prompt/output throughput; exact inference configurations are listed in Appendix~\ref{sec:appendix_systems}.

\paragraph{Open platform.}
We release the full \benchname{} platform, including method adapters,
standardized scripts, inference harnesses, normalized result schemas, and hosted
compressed checkpoints, so that future methods can be compared under the same
model, task, inference, and budget definitions.

\subsection{Evaluation questions}

We organize the empirical study as a causal chain from standardized accuracy claims to deployable efficiency. We first ask whether reported SVD progress survives when protocol differences are removed. We then compare this standardized SVD frontier with structured pruning under the same budget. Finally, we test whether retained accuracy and low-rank parameter savings produce measurable inference gains. These steps define the following questions.

\noindent\textbf{\textsc{Q1.} \emph{Is reported SVD progress robust to standardization?}}
We re-evaluate SVD-style methods under aligned models, tasks, keep-ratio
budgets, recovery settings, and evaluation scripts, isolating the effect of
low-rank subspace selection from protocol differences.

\smallskip
\noindent\textbf{\textsc{Q2.} \emph{Is standardized SVD competitive with structured pruning?}}
Using the same backbones, tasks, and parameter-count budgets, we compare SVD
methods with strong structured pruning baselines to quantify the trade-off
between low-rank factorization and weight removal.

\smallskip
\noindent\textbf{\textsc{Q3.} \emph{Do low-rank savings become inference gains?}}
We export compressed checkpoints through a shared vLLM inference path and measure
prefill latency, decode latency, end-to-end latency, and throughput, testing
whether nominal compression translates into deployable efficiency.

\section{Q1: Re-Assessing Existing SVD Rankings}

\begin{figure}[ht] 
    \centering
    \begin{subfigure}{0.99\linewidth}
        \centering
        \includegraphics[width=\linewidth]{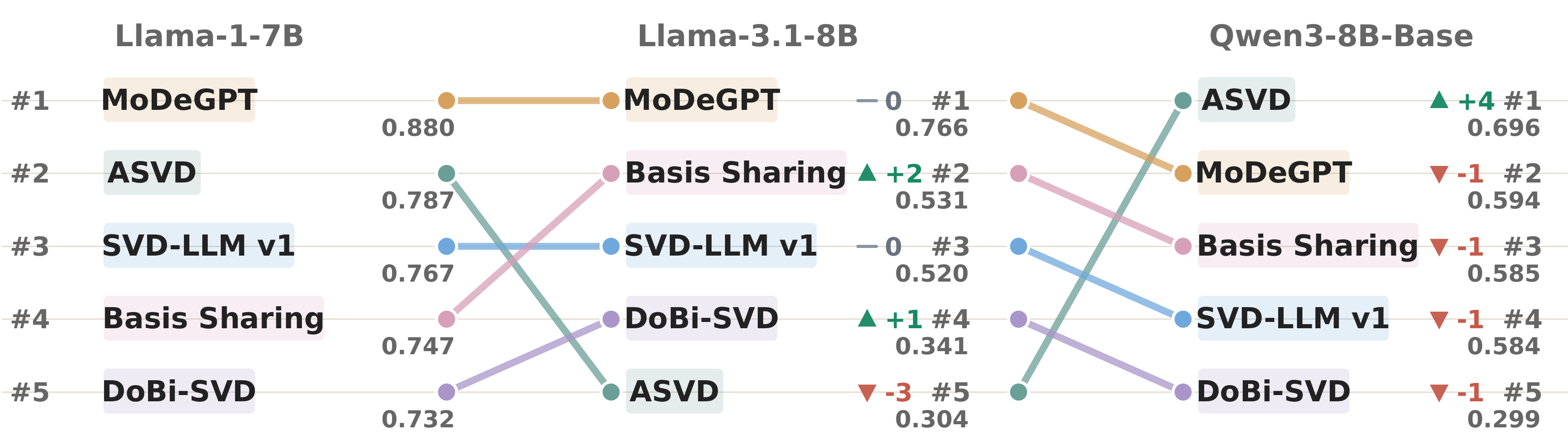}
        \caption{80\% keep ratio.}
        \label{fig:rank_80}
    \end{subfigure}
    
    \vspace{1em} 
    
    \begin{subfigure}{0.99\linewidth}
        \centering
        \includegraphics[width=\linewidth]{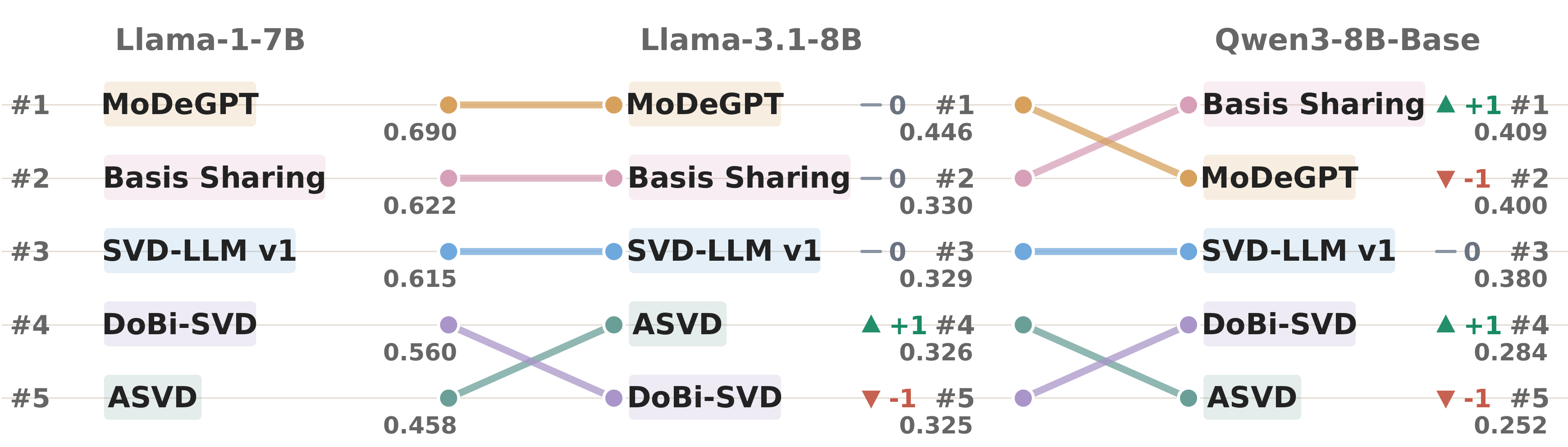}
        \caption{60\% keep ratio.}
        \label{fig:rank_60}
    \end{subfigure}
    
    \caption{\textbf{Method Performance Ranking Shifts Across Model Families.} Method rankings shift substantially when moving from legacy backbones (Llama-1) to modern model families (Llama-3.1 and Qwen3). These shifts show that the apparent superiority of a given SVD-based method is architecture-dependent and does not consistently transfer across model families or keep ratios.}
    \label{fig:rank_transition_all}
\end{figure}

To establish a controlled comparison, we re-evaluate existing SVD methods within our unified framework. Standardizing evaluations through the \textsc{LowRankArena} platform notably reshapes the ranking landscape, revealing that perceived superiority is confounded with setup rather than representing algorithmic merit alone.



\begin{table*}[t]
    \centering
\caption{\textbf{Main standardized leaderboard.}
SVD-style compression methods are evaluated across keep ratios on Llama and Qwen anchor models under matched task versions, evaluation scripts, and budget definitions.
The table highlights that method rankings are not intrinsic: the strongest recipe changes across backbone families, compression budgets, and metric groups.}
    \label{tab:leaderboard_tf}
    \tiny
    \renewcommand{\arraystretch}{1.08}
    \setlength{\tabcolsep}{2.0pt}
    \resizebox{\textwidth}{!}{
    \begin{tabular}{@{}lll!{\vrule width 0.45pt}cc!{\vrule width 0.45pt}ccccccc!{\vrule width 0.45pt}c!{\vrule width 0.45pt}cc!{\vrule width 0.45pt}c@{}}
        \toprule
        & & &
        \multicolumn{2}{c!{\vrule width 0.45pt}}{\textbf{PPL} $\downarrow$} &
        \multicolumn{7}{c!{\vrule width 0.45pt}}{\textbf{General MCQ} $\uparrow$} &
        \multicolumn{1}{c!{\vrule width 0.45pt}}{\textbf{MCQ} $\uparrow$} &
        \multicolumn{2}{c!{\vrule width 0.45pt}}{\textbf{Math} $\uparrow$} &
        \multicolumn{1}{c@{}}{\textbf{Ret.} $\uparrow$} \\
        \cmidrule(lr){4-5}\cmidrule(lr){6-12}\cmidrule(lr){13-13}\cmidrule(lr){14-15}\cmidrule(l){16-16}
        Model & Budget & Method
        & WikiT. & C4
        & BoolQ & ARC-E & ARC-C & WinoG. & PIQA & HellaS. & OBQA
        & Avg.
        & MathQA & MMLU-M
        & Q. Ret. \\
        \midrule

        \rowcolor{gray!10}
        \multirow{12}{*}{Llama-1-7B}
            & \multicolumn{2}{l}{\textbf{Dense FP}}
            & \multicolumn{1}{c}{5.67}
            & \multicolumn{1}{c}{7.20}
            & \multicolumn{1}{c}{0.737}
            & \multicolumn{1}{c}{0.726}
            & \multicolumn{1}{c}{0.445}
            & \multicolumn{1}{c}{0.693}
            & \multicolumn{1}{c}{0.786}
            & \multicolumn{1}{c}{0.749}
            & \multicolumn{1}{c}{0.408}
            & \multicolumn{1}{c}{0.649}
            & \multicolumn{1}{c}{0.261}
            & \multicolumn{1}{c}{0.278}
            & \multicolumn{1}{c@{}}{1.000} \\
        \cmidrule(lr){2-16}

            & \multirow{5}{*}{80\%} & ASVD
            & 8.60 & 11.04
            & \textbf{0.743} & \textbf{0.628} & \textbf{0.388} & \textbf{0.669} & \textbf{0.750} & \textbf{0.696} & 0.392
            & \textbf{0.609}
            & 0.241 & 0.211
            & 0.787 \\
            &  & \textsc{SVD-LLM v1}
            & 7.88 & 16.42
            & 0.633 & 0.481 & 0.317 & 0.584 & 0.681 & 0.477 & 0.332
            & 0.501
            & 0.224 & 0.291
            & 0.767 \\
            &  & DoBi-SVD
            & 9.23 & 19.01
            & 0.620 & 0.500 & 0.305 & 0.635 & 0.665 & 0.570 & 0.385
            & 0.526
            & 0.229 & 0.272
            & 0.732 \\
            &  & Basis Sharing
            & 7.74 & 15.51
            & 0.645 & 0.614 & 0.367 & 0.645 & 0.711 & 0.633 & \textbf{0.400}
            & 0.574
            & 0.239 & 0.205
            & 0.747 \\
            &  & MoDeGPT
            & \textbf{6.92} & \textbf{10.87}
            & 0.647 & 0.587 & 0.353 & 0.637 & 0.724 & 0.582 & 0.336
            & 0.552
            & \textbf{0.254} & \textbf{0.304}
            & \textbf{0.880} \\

        \cmidrule(lr){2-16}

            & \multirow{5}{*}{60\%} & ASVD
            & 3839.83 & 4268.56
            & 0.450 & 0.274 & 0.244 & 0.503 & 0.528 & 0.264 & 0.238
            & 0.357
            & 0.198 & 0.272
            & 0.458 \\
            &  & \textsc{SVD-LLM v1}
            & 13.74 & 55.03
            & 0.380 & 0.392 & 0.258 & 0.538 & 0.572 & 0.350 & 0.304
            & 0.399
            & 0.222 & \textbf{0.296}
            & 0.615 \\
            &  & DoBi-SVD
            & 15.24 & 48.37
            & 0.433 & 0.389 & 0.268 & 0.594 & 0.580 & 0.403 & 0.312
            & 0.426
            & 0.213 & 0.225
            & 0.560 \\
            &  & Basis Sharing
            & \textbf{12.42} & 41.13
            & 0.503 & 0.472 & 0.287 & 0.586 & 0.608 & 0.452 & \textbf{0.344}
            & 0.465
            & 0.209 & 0.268
            & 0.622 \\
            &  & MoDeGPT
            & 12.43 & \textbf{23.40}
            & \textbf{0.613} & \textbf{0.474} & \textbf{0.335} & \textbf{0.645} & \textbf{0.644} & \textbf{0.521} & 0.334
            & \textbf{0.509}
            & \textbf{0.238} & 0.275
            & \textbf{0.690} \\



        \midrule

        \rowcolor{gray!10}
        \multirow{12}{*}{Llama-3.1-8B}
            & \multicolumn{2}{l}{\textbf{Dense FP}}
            & \multicolumn{1}{c}{6.24}
            & \multicolumn{1}{c}{9.10}
            & \multicolumn{1}{c}{0.831}
            & \multicolumn{1}{c}{0.824}
            & \multicolumn{1}{c}{0.549}
            & \multicolumn{1}{c}{0.746}
            & \multicolumn{1}{c}{0.812}
            & \multicolumn{1}{c}{0.793}
            & \multicolumn{1}{c}{0.454}
            & \multicolumn{1}{c}{0.716}
            & \multicolumn{1}{c}{0.396}
            & \multicolumn{1}{c}{0.437}
            & \multicolumn{1}{c@{}}{1.000} \\
        \cmidrule(lr){2-16}

            & \multirow{5}{*}{80\%} & ASVD
            & 2011.38 & 1281.96
            & 0.382 & 0.285 & 0.226 & 0.512 & 0.536 & 0.285 & 0.244
            & 0.353
            & 0.201 & 0.223
            & 0.304 \\
            &  & \textsc{SVD-LLM v1}
            & 14.83 & 80.94
            & \textbf{0.661} & 0.528 & 0.315 & 0.645 & 0.639 & 0.476 & 0.350
            & 0.516
            & 0.256 & 0.305
            & 0.520 \\
            &  & DoBi-SVD
            & 556.59 & 1008.41
            & 0.378 & 0.298 & 0.226 & 0.516 & 0.522 & 0.282 & 0.266
            & 0.355
            & 0.205 & 0.292
            & 0.341 \\
            &  & Basis Sharing
            & 15.61 & 54.36
            & 0.632 & 0.637 & 0.367 & 0.667 & 0.701 & 0.548 & 0.372
            & 0.561
            & 0.248 & 0.297
            & 0.531 \\
            &  & MoDeGPT
            & \textbf{9.01} & \textbf{17.68}
            & 0.412 & \textbf{0.715} & \textbf{0.436} & \textbf{0.730} & \textbf{0.743} & \textbf{0.710} & \textbf{0.382}
            & \textbf{0.590}
            & \textbf{0.344} & \textbf{0.407}
            & \textbf{0.766} \\

        \cmidrule(lr){2-16}

            & \multirow{5}{*}{60\%} & ASVD
            & 22684.63 & 14186.23
            & 0.405 & 0.254 & 0.257 & 0.491 & 0.507 & 0.260 & 0.284
            & 0.351
            & 0.192 & \textbf{0.286}
            & 0.326 \\
            &  & \textsc{SVD-LLM v1}
            & 199.84 & 1187.78
            & 0.378 & 0.295 & 0.246 & 0.533 & 0.515 & 0.283 & 0.268
            & 0.360
            & 0.205 & 0.257
            & 0.329 \\
            &  & DoBi-SVD
            & 987.51 & 1529.38
            & 0.378 & 0.271 & 0.251 & 0.481 & 0.511 & 0.265 & 0.288
            & 0.349
            & 0.203 & 0.268
            & 0.325 \\
            &  & Basis Sharing
            & 82.96 & 461.21
            & 0.380 & 0.409 & 0.241 & 0.562 & 0.568 & 0.325 & 0.284
            & 0.396
            & 0.205 & 0.211
            & 0.330 \\
            &  & MoDeGPT
            & \textbf{24.50} & \textbf{51.82}
            & \textbf{0.622} & \textbf{0.460} & \textbf{0.312} & \textbf{0.672} & \textbf{0.629} & \textbf{0.516} & \textbf{0.316}
            & \textbf{0.504}
            & \textbf{0.241} & 0.213
            & \textbf{0.446} \\



                \midrule

        \rowcolor{gray!10}
        \multirow{12}{*}{Qwen3-8B-Base}
            & \multicolumn{2}{l}{\textbf{Dense FP}}
            & \multicolumn{1}{c}{7.00}
            & \multicolumn{1}{c}{11.78}
            & \multicolumn{1}{c}{0.830}
            & \multicolumn{1}{c}{0.800}
            & \multicolumn{1}{c}{0.570}
            & \multicolumn{1}{c}{0.727}
            & \multicolumn{1}{c}{0.793}
            & \multicolumn{1}{c}{0.787}
            & \multicolumn{1}{c}{0.420}
            & \multicolumn{1}{c}{0.704}
            & \multicolumn{1}{c}{0.542}
            & \multicolumn{1}{c}{0.729}
            & \multicolumn{1}{c@{}}{1.000} \\
        \cmidrule(lr){2-16}

            & \multirow{5}{*}{80\%} & ASVD
            & 11.88 & \textbf{20.54}
            & \textbf{0.792} & \textbf{0.758} & \textbf{0.483} & 0.651 & \textbf{0.745} & 0.641 & \textbf{0.414}
            & \textbf{0.641}
            & \textbf{0.420} & \textbf{0.460}
            & \textbf{0.696} \\
            &  & \textsc{SVD-LLM v1}
            & 11.08 & 33.90
            & 0.688 & 0.649 & 0.424 & 0.669 & 0.712 & 0.616 & 0.408
            & 0.595
            & 0.330 & 0.355
            & 0.584 \\
            &  & DoBi-SVD
            & 51.21 & 236.65
            & 0.460 & 0.335 & 0.284 & 0.519 & 0.550 & 0.388 & 0.262
            & 0.400
            & 0.206 & 0.264
            & 0.299 \\
            &  & Basis Sharing
            & 11.05 & 31.67
            & 0.676 & 0.578 & 0.439 & \textbf{0.673} & 0.726 & 0.635 & 0.388
            & 0.588
            & 0.331 & 0.347
            & 0.585 \\
            &  & MoDeGPT
            & \textbf{10.34} & 22.46
            & 0.658 & 0.613 & 0.430 & 0.655 & 0.731 & \textbf{0.683} & 0.408
            & 0.597
            & 0.279 & 0.297
            & 0.594 \\

        \cmidrule(lr){2-16}

            & \multirow{5}{*}{60\%} & ASVD
            & 1359.38 & 1484.57
            & 0.502 & 0.292 & 0.241 & 0.499 & 0.544 & 0.281 & 0.270
            & 0.375
            & 0.212 & 0.234
            & 0.252 \\
            &  & \textsc{SVD-LLM v1}
            & 20.44 & 112.01
            & 0.544 & 0.382 & 0.265 & 0.545 & 0.590 & 0.385 & 0.266
            & 0.425
            & 0.231 & \textbf{0.307}
            & 0.380 \\
            &  & DoBi-SVD
            & 107.63 & 610.21
            & 0.511 & 0.290 & 0.253 & 0.511 & 0.523 & 0.310 & 0.306
            & 0.386
            & 0.203 & 0.302
            & 0.284 \\
            &  & Basis Sharing
            & 18.59 & 95.39
            & \textbf{0.630} & \textbf{0.457} & 0.278 & \textbf{0.580} & 0.620 & 0.423 & 0.298
            & \textbf{0.469}
            & \textbf{0.253} & 0.302
            & \textbf{0.409} \\
            &  & MoDeGPT
            & \textbf{18.40} & \textbf{55.66}
            & 0.509 & 0.414 & \textbf{0.293} & 0.569 & \textbf{0.625} & \textbf{0.460} & \textbf{0.328}
            & 0.457
            & 0.228 & 0.246
            & 0.400 \\
        
        \bottomrule
    \end{tabular}}

\end{table*}


\paragraph{Context-Dependent Rankings.} Table~\ref{tab:leaderboard_tf} shows that rankings are context-dependent rather than universal, and the per-task columns explain why. On Llama backbones, MoDeGPT leads in aggregate retention at moderate ratios by avoiding the severe perplexity collapse seen in other training-free baselines. For example, on Llama-3.1-8B at 60\% keep, SVD-LLM and Basis Sharing reach C4 perplexities of 1187.78 and 461.21, while MoDeGPT remains stable at 51.82. However, this advantage is not absolute; as ratios drop, individual reasoning or math metrics often favor different methods. This implies the aggregate score reflects a trade-off between generative stability and downstream accuracy, rather than a universal ordering of method quality. The anomalous 0.41 BoolQ score was exactly reproduced for the same checkpoint and reflects a strong output-label bias; its sensitivity to calibration draws is analyzed in Appendix~\ref{app:boolq_audit}. 

%



\paragraph{Architectural Sensitivity.} Figure~\ref{fig:rank_transition_all} illustrates this instability by tracking quality-retention ranks across model families. While MoDeGPT leads on Llama backbones at 80\% and 60\% keep, lower rankings shift substantially: ASVD drops from \#2 to \#5 at 80\% keep when moving from Llama-1 to Llama-3.1, while Basis Sharing rises from \#4 to \#2. Qwen3-8B-Base results confirm that even the leader is not architecture-invariant; ASVD and Basis Sharing take the top spots at 80\% and 60\% keep, respectively, pushing MoDeGPT to second. These transitions reinforce the need for standardization: the "best" SVD recipe varies with the backbone, budget, and metric mix, necessitating evaluation within a unified platform like \textsc{LowRankArena}.









\begin{figure}[ht]
    \centering
    \includegraphics[width=0.99\linewidth]{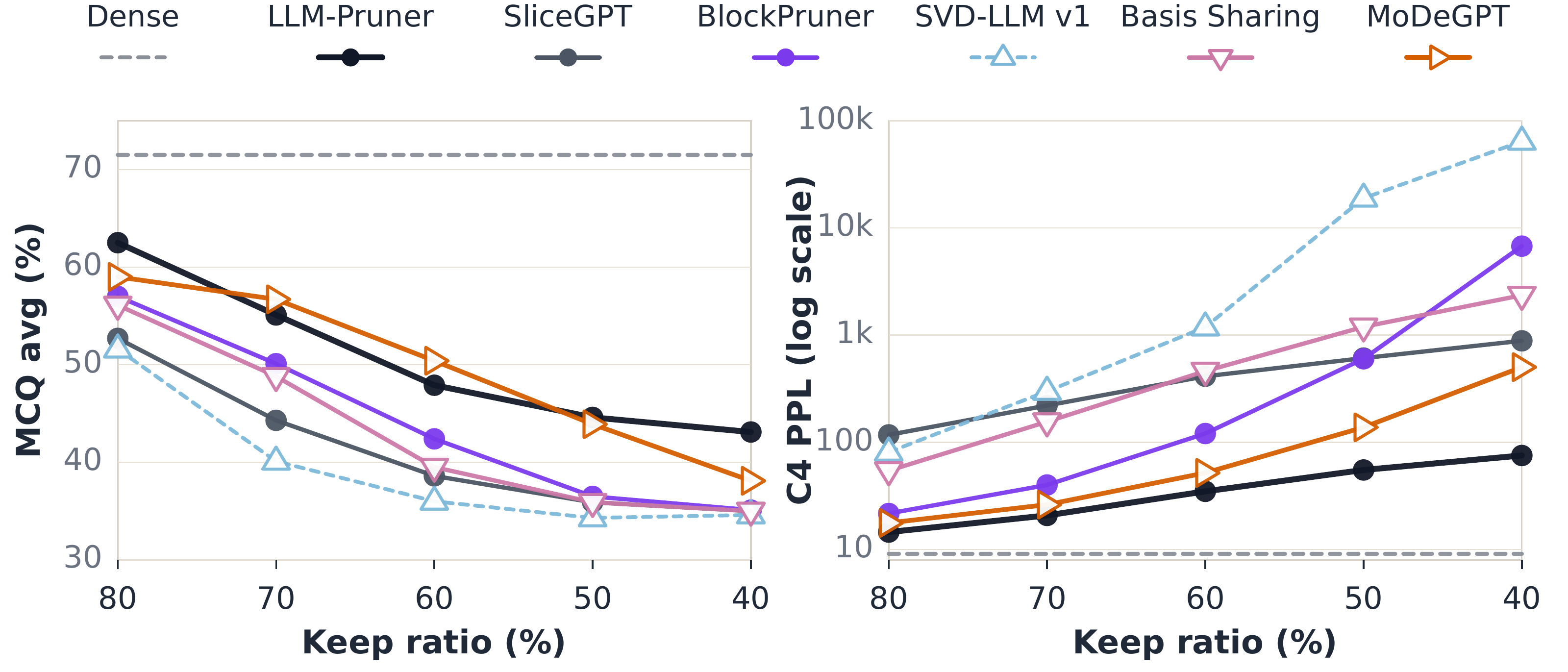}
    \caption{\textbf{The capability cliff in zero-shot low-rank compression.} SVD vs. structured pruning on Llama-3.1-8B under matched ratios (no added post-compression recovery, uniform-choice macro floor 35.7\%). \textbf{Left:} SVD remains competitive with pruning in reasoning accuracy (MCQ Avg). \textbf{Right:} Several SVD methods show severe language-modeling degradation in C4 PPL at aggressive keep ratios, while pruning degrades more gradually.}
    
    \label{fig:capacity_vs_pruning}
\end{figure}

\section{Q2: Comparison with Structured Pruning}

The gap between SVD compression and structured pruning becomes clearer when evaluation moves beyond multiple-choice accuracy. As shown in Figure~\ref{fig:capacity_vs_pruning}, current SVD methods remain competitive with pruning baselines on MCQ benchmarks, but C4 perplexity exposes a larger gap in language-modeling quality. This discrepancy is only partially floor-driven. ASVD’s 0.353 MCQ average sits at the 0.357 macro choice floor and should not be interpreted as retained capability. In contrast, \textsc{SVD-LLM} and Basis Sharing remain above chance across all seven tasks despite elevated C4 perplexities of 80.94 and 54.36.

On Llama-3.1-8B, the strongest pruning baseline degrades more gracefully across keep ratios, whereas several training-free SVD methods encounter a capability cliff under aggressive compression. At the 60\% keep ratio, for example, \textsc{SVD-LLM} and Basis Sharing reach 1187.78 and 461.21 C4 PPL, respectively, while MoDeGPT remains much more stable at 51.82. Even MoDeGPT, however, fails to close the gap to LLM-Pruner, which achieves 34.85 C4 PPL under the same budget.



Thus, SVD methods remain competitive with structured pruning on downstream accuracy, though without absolute dominance across all regimes. The perplexity gap at aggressive ratios highlights room for improving zero-shot SVD, motivating generative stability as a core evaluation metric. Finally, unmatched method default recovery protocols across pruning and low-rank prevent claiming that the ranking in Figure~\ref{fig:capacity_vs_pruning} persists after post-compression recovery.


\section{Q3: End-to-End Inference Gains}
\label{sec:serving_quests}

The translation from low-rank savings to inference gains becomes \textbf{\emph{much less clear once evaluation moves from nominal computation to end-to-end generation}}. Although factorized weights reduce parameter count and theoretical FLOPs, these savings mainly benefit compute-bound stages such as prefill. Decode-heavy workloads are often bottlenecked by memory bandwidth, kernel launch overhead, and the inefficiency of executing dual low-rank GEMMs compared to a single optimized dense GEMM~\cite{shao2026flashsvd}. To answer Q3, we evaluate representative prefill- and decode-heavy workloads under a matched vLLM-based~\cite{kwon2023efficient} inference setup with fixed hardware, prompt profiles, and request policies.

\begin{figure}[ht]
    \centering
    \includegraphics[width=0.99\linewidth]{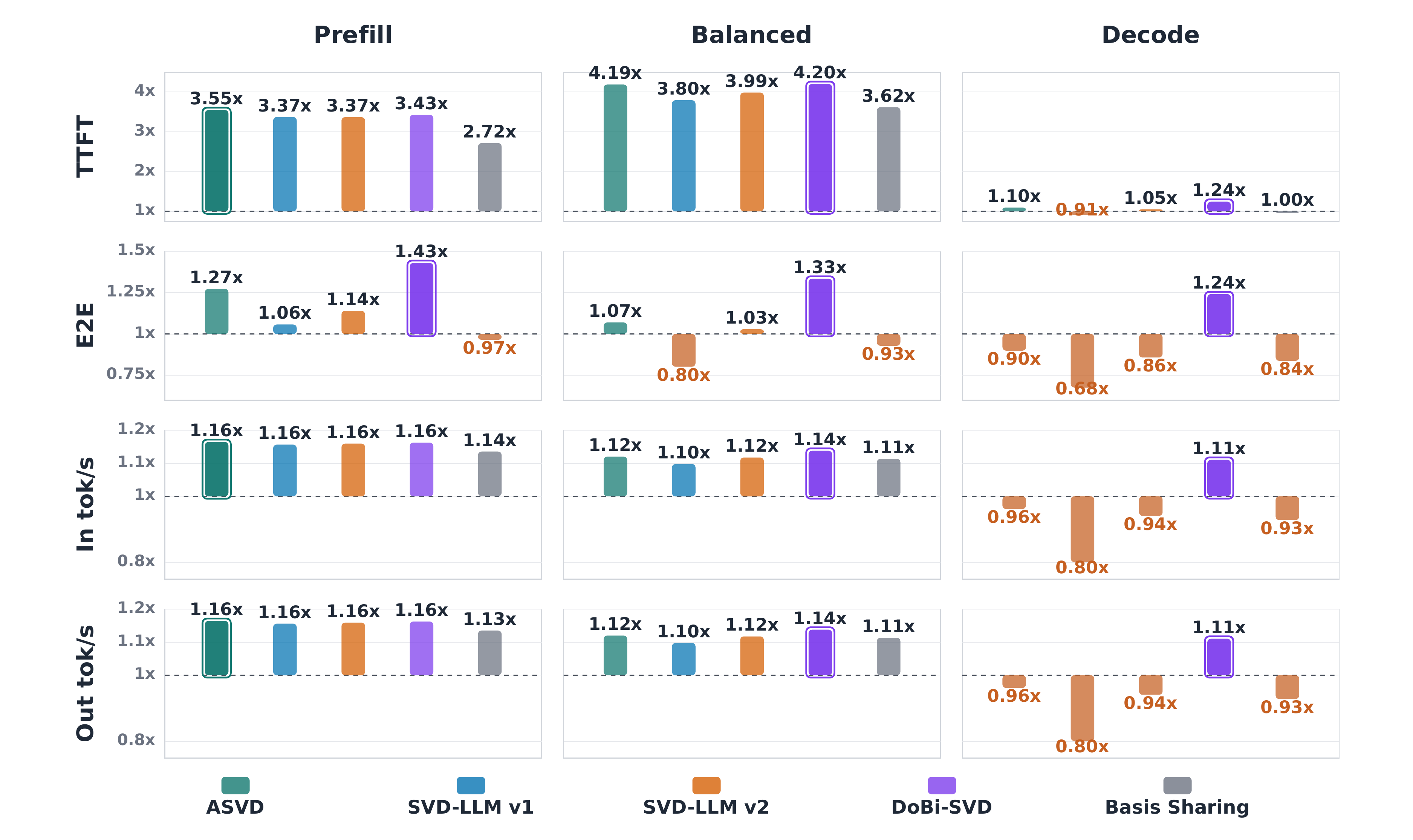}
    \caption{\textbf{Inference speedup across diverse request profiles.}
    End-to-end gains are evaluated across the representative workloads defined
    in Sec.~\ref{sec:appendix_systems}. While SVD reduces nominal computation,
    empirical throughput gains remain marginal and inconsistent, especially
    during decode-heavy generation.}
    \label{fig:serving_summary}
\end{figure}


Figure~\ref{fig:serving_summary} reveals a stark divergence between compute-bound and memory-bound performance. In prefill-heavy and balanced workloads, SVD methods achieve significant TTFT speedups (e.g., up to 4.20x), as reduced arithmetic directly accelerates prompt processing. However, these gains diminish sharply during decoding; in decode-heavy scenarios, most methods yield marginal or even sub-1$\times$ throughput improvements--with SVD-LLM dropping to 0.80$\times$. This confirms that once the bottleneck shifts to memory traffic and kernel launch overhead, the theoretical savings of low-rank factorization fail to translate into practical end-to-end efficiency.

A caveat is that not every speedup reflects pure factorized low-rank execution. DoBi-SVD can fall back to a dense execution path for high-rank layers, allowing the backend to use an optimized dense GEMM instead of two smaller low-rank GEMMs. This hybrid runtime policy can improve latency, especially in decode-heavy settings, but it also means the measured speedup is not attributable to low-rank factorization alone.

\section{Secondary Audits, Scope, and Limitations}
\label{sec:secondary_audits}

Beyond the core experiments, the following audits delineate the operational boundaries of \textsc{LowRankArena}. We examine how conclusions depend on specific artifacts, resource budgets, and deployment environments, rather than introducing additional method rankings.

\noindent
\begin{minipage}[t]{0.52\textwidth}
\vspace{0pt}
\paragraph{Calibration Sensitivity and Ranking Uncertainty.}
The primary leaderboard retains each method's released calibration recipe; calibration inputs are therefore recorded but not fully matched in Table~\ref{tab:leaderboard_tf}. To quantify how this choice affects the performance comparisons in Fig.~\ref{fig:rank_transition_all}, we conduct a controlled audit on Llama-3.1-8B at an 80\% keep ratio. For each calibration corpus, all methods receive the same ordered tensor consisting of 256 sequences, each with 2,048 tokens constructed with a pinned tokenizer.
\end{minipage}%
\hfill%
\begin{minipage}[t]{0.45\textwidth}
\vspace{0pt}
\makeatletter\def\@captype{table}\makeatother
\caption{\small Calibration sensitivity on Llama-3.1-8B at an 80\% keep ratio ($R_Q$, higher is better). WikiText-2 uses 3 resamples.}
\label{tab:calibration_sensitivity}
\small
\setlength{\tabcolsep}{2pt}
\begin{tabular}{lccc}
    \toprule
    \textbf{Method} & \textbf{WikiText-2} & \textbf{C4} & \textbf{Pile-Val} \\
    \midrule
    MoDeGPT       & $0.7699 \pm 0.0039$ & 0.7782 & 0.7801 \\
    Basis Sharing & $0.5369 \pm 0.0037$ & 0.4980 & 0.5150 \\
    SVD-LLM       & $0.5212 \pm 0.0025$ & 0.5033 & 0.4919 \\
    ASVD          & $0.3278 \pm 0.0145$ & 0.3033 & 0.3320 \\
    \bottomrule
\end{tabular}
\end{minipage}

We vary only the calibration corpus or, for WikiText-2, the sampling key, while keeping all remaining compression and evaluation settings fixed. 

The performance ordering remains unchanged across the three WikiText-2 resamples. Changing the corpus shifts absolute retention and reverses the close Basis Sharing--SVD-LLM pair under C4, while MoDeGPT remains the clear leader in every condition. The largest within-WikiText-2 range is $0.0284$, below the 80\%-keep leader margins in Fig.~\ref{fig:rank_transition_all} ($0.093$--$0.235$), although it is not a universal uncertainty bound. These results support architecture-dependent clear leaders and performance tiers, while smaller ordering differences should be treated as near-ties. DoBi-SVD is omitted because its repeated compression runs did not complete within the available compute budget.

\paragraph{Instruction-Tuning and Extended Backbones.} 
To probe the robustness and scope of our findings, we conduct additional audits on instruction-tuned models and expanded backbone variants (e.g., Llama-2, Qwen3). These extensive sweeps, detailed in Appendices~B.4 and B.5, reinforce Q1 by showing that rankings remain strictly conditional on keep ratios across generations. Notably, these results reveal a decoupling effect where downstream MCQ accuracy persists even as perplexity degrades sharply. We categorize instruction-tuned performance as auxiliary evidence rather than a primary claim, as its heightened sensitivity to prompt templates and decoding configurations further justifies our focus on base-model standardization.

\begin{wraptable}{r}{0.5\textwidth}
    \centering
    \vspace{-1.2em}
    \caption{\textbf{Fixed-setting speedup audit.} Llama-3.1-8B, \textsc{SVD-LLM} v1, 0.6 keep ratio on RTX A5000.}
    \label{tab:svdllm_keep06_serving_sanity}
    \renewcommand{\arraystretch}{1.2} 
    \resizebox{\linewidth}{!}{
        \begin{tabular}{lcc}
        \toprule
        Profile & TTFT speedup & E2E speedup \\
        \midrule
        Prefill-heavy & \makecell{\textbf{3.37$\times$} \\ \scriptsize ($3417\!\to\!1013$\,ms)} & \makecell{\textbf{1.06$\times$} \\ \scriptsize ($4925\!\to\!4659$\,ms)} \\
        Balanced      & \makecell{\textbf{3.80$\times$} \\ \scriptsize ($1668\!\to\!439$\,ms)} & \makecell{\textbf{0.80$\times$} \\ \scriptsize ($5905\!\to\!7358$\,ms)} \\
        Decode-heavy  & \makecell{\textbf{0.91$\times$} \\ \scriptsize ($187\!\to\!205$\,ms)} & \makecell{\textbf{0.68$\times$} \\ \scriptsize ($15136\!\to\!22359$\,ms)} \\
        \bottomrule
        \end{tabular}
    }
    \vspace{-1.2em}
\end{wraptable}
\paragraph{Inference on More Devices.}
Cross-device reruns reinforce Q3: nominal low-rank savings do not guarantee stable endpoint speedups. As shown in Table~\ref{tab:svdllm_keep06_serving_sanity}, even at an aggressive 0.6 ratio, large TTFT gains in prefill-heavy settings translate to only marginal E2E improvements and vanish once decoding dominates. Speedups remain highly sensitive to workload phase, GPU type, and kernel execution paths. These audits define the scope of \textsc{LowRankArena} as a platform for standardized comparison under aligned scripts and budgets, rather than an exhaustive search for custom kernel optimizations.

\paragraph{Implementation Readiness under a Fixed Budget.}
We define feasibility through the lens of implementation readiness: the ability of a released codebase to produce standardized \textsc{LowRankArena} artifacts within a fixed one-H200 recipe. This operational focus separates a method's theoretical scalability from its current out-of-the-box utility. Readiness varies significantly across the field: while several methods expose model-generic entry points through Hugging Face auto-loaders or family-level adapters, the released implementations often remain tightly coupled to the dependency versions, numerical kernels, offload assumptions, and architecture patterns in their original studies.

Scaling to 70B exposes these gaps sharply. Under the one-H200 audit, failures stem from code bottlenecks rather than algorithmic limits: ASVD is blocked by a legacy \texttt{lm\_eval} dependency; \textsc{SVD-LLM} v1 and DoBi-SVD hit memory ceilings; Basis Sharing crash in CUDA eigensolvers; and MoDeGPT fails during Accelerate GPU-offloading. Consequently, this audit does not rank intrinsic algorithmic potential. Instead, it records engineering maturity, identifying where dependency repairs, numerical fallbacks, or sharding are required for community-ready 70B generation.

\section{Related Works}

\paragraph{SVD Compression for LLMs.} Low-rank compression for LLMs has evolved from early encoder-based weighting methods~\cite{chen2021drone,hsu2022fwsvd} to a diverse set of decoder-oriented approaches. Recent methods typically improve low-rank recovery through activation-aware rescaling (e.g., ASVD, SVD-LLM~\cite{yuan2023asvd,wang2025svdllm}), loss-sensitive truncation (e.g., DoBi-SVD, \textsc{Zssvd}~\cite{wang2025dobi,abbasi2026zssvd}), parameter sharing (i.e., Basis Sharing, D-Rank~\cite{mi2025drank,wang2025basis}), or adaptive rank allocation~\cite{gao2026dfsvd,lin2024modegpt,li2025adasvd}. Despite this progress, these methods are often evaluated under heterogeneous experimental setups, making reported results difficult to compare directly. \textsc{LowRankArena} addresses this gap by re-evaluating these families under a standardized protocol with aligned budgets, tasks, artifacts, and inference measurements.


\paragraph{Attention-Side Low-Rank and Runtime Systems.} Methods such as Palu, xKV, and QSVD~\cite{chang2025palu,chang2025xkv,saxena2024eigen,wang2025qsvd} target attention-side memory for long-context serving. We treat these as complementary, as \textsc{LowRankArena} focuses on static checkpoint compression, which is the primary constraint in edge or short-context deployment. Furthermore, system works like vLLM and FlashSVD~\cite{kwon2023efficient,shao2026flashsvd} show that practical speed depends on memory movement and backend execution rather than FLOP reduction alone, justifying our focus on end-to-end serving audits to capture real-world kernel impact.


\paragraph{Structured Pruning and Hybrid Schemes.} Structured pruning removes architectural components (e.g., heads or blocks) to produce smaller dense models~\cite{ma2023llm,ashkboos2024slicegpt,zhong2025blockpruner}, whereas SVD factorization preserves original module interfaces. \textsc{LowRankArena} treats state-of-the-art pruning as a critical deployment-level baseline under matched parameter budgets. While quantization and hybrid low-rank methods~\cite{abbasi2026zssvd} offer orthogonal compression axes, mixing rank reduction with precision changes can obscures the isolated effect of low-rank subspace selection. We therefore focus our primary benchmarks on uniform-precision SVD and treat hybrid variants as auxiliary analysis.

\paragraph{LLM Evaluation and Compression Benchmarks.}
General-purpose evaluation tools such as LM-Eval-Harness provide broad task coverage and reproducible task execution~\cite{biderman2024lm_eval,eleutherai2026lm_eval_v0411}, while compression benchmarks such as LLMCBench highlight the need to evaluate compression methods across models, metrics, and deployment constraints~\cite{yang2024llmcbench}. However, existing infrastructure is not specialized for SVD checkpoint compression, where calibration data, keep-ratio semantics, task versions, artifacts, and inference backends often differ across papers. \textsc{LowRankArena} complements these efforts by fixing the budget axis, aligning task and inference setups, separating pure low-rank compression from auxiliary techniques, and releasing standardized checkpoints for reproducible future comparison.

\section{Conclusion}


This paper presents \textsc{LowRankArena}, a unified platform for standardized auditing of SVD-based LLM compression. By enforcing evaluation standardization across tasks and backbones, we show that reported progress often becomes more conditional under aligned protocols. Our findings reveal no method remains the clear leader across all evaluated backbones; clear leaders and performance tiers depend on the backbone and budget, while close orderings can be calibration-sensitive. Furthermore, empirical efficiency remains highly workload-dependent, with speedups concentrated in compute-bound prefill stages rather than the full generation lifecycle. \textsc{LowRankArena} thus provides a rigorous foundation for reproducible, protocol-aligned comparison of future SVD compression methods.



\bibliographystyle{plain}
\bibliography{text/refs} 

\newpage
\appendix

\section*{Organization of the Appendix}
In this Appendix, we provide additional technical background, platform implementation details, and extended experimental audits that complement the main paper. The document is organized as follows:

\begin{itemize}
    \item \textbf{Section \ref{app:setups}: Experimental Setups and Evaluation Suites}
    \begin{itemize}
        \item Detailed task and metric configurations for base and instruction-tuned suites.
        \item System specifications for the \textsc{LowRankArena} inference harness and vLLM backends.
        \item Statistical rigor: Error bars, request counts, and latency profiling protocols.
    \end{itemize}

    \item \textbf{Section \ref{app:field_audit}: Extended Field Auditing and Robustness}
    \begin{itemize}
        \item \textbf{Prior Evidence Audit:} A systematic review of speed claims in existing SVD literature.
        \item \textbf{Cross-Device Robustness:} Expanded inference benchmarks across A100, L40S, and RTX A5000.
        \item \textbf{Feasibility and Cost:} Large-model (70B) compression feasibility and wall-clock cost breakdown.
        \item \textbf{Instruction-Tuning Sweeps:} Detailed rankings and sensitivity analysis for chat-style models.
        \item \textbf{BoolQ Label-Bias Audit:} Empirical analysis of output-label bias anomalies and their sensitivity to calibration draws.
    \end{itemize}

    \item \textbf{Section \ref{app:platform_impl}: Platform Architecture and Artifact Management}
    \begin{itemize}
        \item \textbf{Reproducible Pipeline:} Design of the YAML-driven orchestration and standardized SVD logic.
        \item \textbf{Artifact Documentation:} Metadata and organization of the released 3+ TiB checkpoint zoo with artifact documentation.
        \item \textbf{Standardization Protocols:} How we enforce uniform-precision and budget alignment across diverse methods.
    \end{itemize}
   
    \item \textbf{Section \ref{app:limits}: Discussion, Limitations, and Maintenance Plan}
    \begin{itemize}
        \item Scope of current SVD kernels and future extension to quantization/hybrid methods.
        \item Maintenance, hosting, and community contribution guidelines for the benchmark.
    \end{itemize}
\end{itemize}

\section{Setups}
\label{app:setups}

\subsection{Task and Metric Configuration}
\label{sec:appendix_tasks}

All accuracy and generation benchmarks are run with LM-Eval-Harness v0.4.11~\cite{eleutherai2026lm_eval_v0411} unless otherwise stated. LowRankArena evaluates base/pretrained checkpoints and instruction-tuned checkpoints with separate suites. Base models are not evaluated with chat templates or free-form instruction-following protocols. Publication years of selected methods are ASVD (2023); SVD-LLM, Basis Sharing, and MoDeGPT (2024); and DoBi-SVD (2025).

\paragraph{Base model suite.}
The base suite contains three parts:
\[
\texttt{base} =
\texttt{ppl} + \texttt{mcq} + \texttt{base/base\_math}.
\]

The \texttt{mcq} suite is evaluated in the 0-shot setting. It includes \texttt{boolq}~\cite{clark2019boolq}, \texttt{arc\_easy}~\cite{clark2018arc}, \texttt{arc\_challenge}~\cite{clark2018arc}, \texttt{winogrande}~\cite{sakaguchi2021winogrande}, \texttt{piqa}~\cite{bisk2020piqa}, \texttt{hellaswag}~\cite{zellers2019hellaswag}, and \texttt{openbookqa}~\cite{mihaylov2018can}. The headline metric is \texttt{acc\_norm}. When normalized accuracy is not available, we use \texttt{acc}. We define a \emph{task version} by the complete evaluation specification: the harness version, task alias and split, prompt and few-shot configuration, preprocessing, and metric. We report the macro average across tasks. 

Perplexity is measured with LowRankArena's contiguous-text runner instead of LM-Eval-Harness rolling windows. We tokenize raw text without chat templates or special-token insertion. The tokenized text is split into non-overlapping 2048-token blocks. We report perplexity on WikiText-2 test and a fixed C4 validation stream.


The \texttt{base/base\_math} suite evaluates base-model math retention with multiple-choice or log-likelihood scoring. It combines 0-shot \texttt{lra\_mathqa}~\cite{amini2019mathqa} with 5-shot \texttt{MMLU\_Math}~\cite{hendrycks2021mmlu}. The local \texttt{lra\_mathqa} wrapper uses the official MathQA data and avoids dataset-loader incompatibilities in the pinned environment.

\paragraph{Instruction-tuned suite.}
The main instruction-tuned suite contains MMLU-Pro and GSM8K:
\[
\texttt{instruct} =
\texttt{instruct/mmlu\_pro} + \texttt{instruct/gsm8k}.
\]

MMLU-Pro uses the upstream \texttt{leaderboard\_mmlu\_pro}~\cite{wang2024mmlupro} task with
5-shot direct-answer multiple-choice scoring and reports \texttt{acc}.
GSM8K uses \texttt{gsm8k\_cot} with 8-shot chain-of-thought prompting and
reports \texttt{exact\_match}. We do not use self-consistency.


We also report one appendix instruction-tuned task:
\[
\texttt{instruct\_appendix} = \texttt{instruct/ifeval}.
\]
IFEval uses the upstream \texttt{ifeval} task with the
tokenizer chat template enabled. Its headline metric is
\texttt{prompt\_level\_strict\_acc}. We also retain the strict and loose
prompt-level and instruction-level accuracies in the raw results.

\paragraph{Calibration and recovery protocol.}
For Table~\ref{tab:leaderboard_tf} and Fig.~\ref{fig:rank_transition_all}, all calibration-based methods use WikiText-2 sequences of length 2,048: ASVD uses 32 sequences, MoDeGPT uses 128, and SVD-LLM, DoBi-SVD, and Basis Sharing each use 256; we use the native variant of each method except \texttt{whitening\_only} for SVD-LLM, while pruning uses \texttt{prune\_only} without calibration, and no method receives added recovery (\texttt{update\_u256} and \texttt{alpaca\_recover} are excluded).

\paragraph{Aggregation and backend.}
For each multi-task suite, LowRankArena reports the macro mean of the
configured headline metrics. Accuracy suites use \texttt{dtype=auto},
\texttt{batch\_size=auto}, and the vLLM LM-Eval backend by default.
Perplexity uses the contiguous-text runner described above. The released
benchmark YAML files contain the full task aliases, prompt settings, and
runner options.

\subsection{System and Inference Configuration}
\label{sec:appendix_systems}

\textsc{LowRankArena} reports inference measurements from its end-to-end inference harness rather than layer-level microbenchmarks. For each checkpoint, the harness starts a fresh vLLM OpenAI-compatible server, runs \texttt{vllm bench serve}, records the latency and throughput outputs, and then terminates the server before evaluating the next checkpoint.

\paragraph{Hardware and run provenance.}
The primary inference benchmark utilizes one NVIDIA A100 80GB PCIe GPU, 12 CPU cores, and 192 GB of host memory. To validate cross-device robustness, we replicate these evaluations on RTX A5000 and L40S hardware. Additionally, large-scale (70B) feasibility bounds are audited on a single H200 141GB GPU. When a table reports a specific hardware rerun, the device and precision are fixed as table-level configurations rather than method variables. For instance, the BF16 Llama-3.1-8B rerun on the RTX A5000 uses vLLM \texttt{0.18.1}, tensor-parallel size 1, \texttt{max\_model\_len=4608}, request rate 1.0, 64 prompts per profile, and maximum concurrency 16. Separate A100-40 pruning feasibility runs employ a distinct stress-style configuration: \texttt{bfloat16}, \texttt{request\_rate=inf}, \texttt{num\_prompts=3}, \texttt{max\_concurrency=1}, and \texttt{gpu\_memory\_utilization=0.75}.

\paragraph{Online inference protocol and profiles.}
The primary online protocol relies on the \texttt{speed/serve\_e2e} suite. To strictly isolate the effects of low-rank compression, all methods within a given result table share the exact same request stream, vLLM scheduler, tokenizer path, tensor-parallel degree, dtype, and KV-cache policy (detailed in Table~\ref{tab:systems_serving_config}). Checkpoint preparation is limited to standard vLLM loading logic; we do not employ method-specific batching rules, request policies, or custom kernels.

The suite launches \texttt{vllm serve} and issues synthetic random-token requests via the \texttt{/v1/completions} endpoint. This avoids prompt-content variability while keeping requested prompt and generation lengths fixed. Generation uses temperature \(0\), disables prompt shuffling, and ignores EOS. Each method is evaluated across three standardized profiles (Table~\ref{tab:systems_serving_profiles}): prefill-heavy, balanced, and decode-heavy. We record time-to-first-token (TTFT), time per output token (TPOT), inter-token latency (ITL), end-to-end latency, and token throughput. Unless stated otherwise, tables report p50 latency and mean throughput, while raw JSON artifacts retain p95 and p99 distributions.

\paragraph{Offline throughput sanity checks.}
\textsc{LowRankArena} also includes an in-process vLLM throughput suite, \texttt{speed/serve}, to verify prefill, decode, and end-to-end tokens per second without HTTP server overhead. This suite calls \texttt{LLM.generate} directly with deterministic repeated-token prompts. It uses one warmup iteration, seven measured repeats, \texttt{temperature=0}, and \texttt{ignore\_eos=true}. The standard offline cases cover batch sizes \(\{1,2,4\}\), prompt lengths \(\{512,1024,2048\}\), and generation lengths \(\{128,512\}\). We use \texttt{speed/serve\_e2e} for all primary endpoint latency claims, reserving \texttt{speed/serve} strictly for implementation and sanity checks.

\begin{table}[t]
\centering
\small
\caption{\textbf{Inference configuration.}
Primary LowRankArena online inference settings are taken from
\texttt{benchmark/speed/serve\_e2e.yaml} and the corresponding
inference-summary launch scripts.}
\label{tab:systems_serving_config}
\begin{tabular}{p{0.31\linewidth}p{0.61\linewidth}}
\toprule
Item & Configuration \\
\midrule
Benchmark suite &
LowRankArena \texttt{speed/serve\_e2e}. \\

Inference backend &
\texttt{vllm serve}; backend version recorded as vLLM \texttt{0.18.1}
in the normalized outputs. \\

Client driver &
\texttt{vllm bench serve} with \texttt{--backend openai} and
\texttt{--endpoint /v1/completions}. \\

Tensor parallelism &
\texttt{tensor\_parallel\_size=1}; one GPU per server. \\

Precision &
Fixed within each table. The A100 launch script uses \texttt{float16}.
BF16 reruns set \texttt{bfloat16} for every method in the same table. \\

KV cache and scheduler &
vLLM defaults for the chosen dtype and engine settings; no
method-specific scheduler options. \\

Engine memory cap &
Default \texttt{gpu\_memory\_utilization=0.6} for the A100 launch script.
Rerun overrides are reported with the result set. \\

Maximum model length &
\texttt{max\_model\_len=4608} for the three standard profiles. \\

Eager mode &
\texttt{enforce\_eager=false} unless a smoke or debugging run explicitly
overrides it. \\

Trust remote code &
\texttt{true}, needed for exported low-rank wrappers and custom checkpoint
layouts. \\

Request arrival &
Open-loop \texttt{request\_rate=1.0} request/s in the standard
inference-summary suite. \\

Requests per profile &
\texttt{num\_prompts=64} and \texttt{max\_concurrency=16} in the standard
inference-summary suite. \\

Warmup and seed &
\texttt{num\_warmups=1}, \texttt{seed=0}, and
\texttt{random\_range\_ratio=0.0} in the YAML default. Table-specific
reruns may increase warmups, but the value is kept the same for all methods
in that table. \\

Latency percentiles &
p50, p95, and p99 for TTFT, TPOT, ITL, and end-to-end latency. \\
\bottomrule
\end{tabular}
\end{table}

\begin{table}[t]
\centering
\small
\caption{\textbf{Inference profiles.}
Each method in a inference table is evaluated on the same prefill-heavy,
balanced, and decode-heavy profiles with the same request count, request
arrival process, and engine settings.}
\label{tab:systems_serving_profiles}
\resizebox{\textwidth}{!}{
\begin{tabular}{lrrl}
\toprule
Profile & Input tokens & Output tokens & Purpose \\
\midrule
Prefill-heavy & 4096 & 32 &
Measures prompt processing with a long input and short output. \\
Balanced & 2048 & 128 &
Measures a mixed prompt-processing and generation setting. \\
Decode-heavy & 512 & 512 &
Measures autoregressive generation with a long output. \\
\bottomrule
\end{tabular}
}
\end{table}

\section{Field Auditing}
\label{app:field_audit}
\subsection{Prior Speed Evidence}

\paragraph{Prior SVD speed evidence.}
Tables~\ref{tab:svd_compact_maintext_a} and~\ref{tab:svd_compact_maintext_b} summarize representative SVD-style compression papers and separate the core compression idea from the reported speed evidence. The trend is that several works include real hardware measurements, but the protocols often differ in device, workload shape, sequence length, batch size, or endpoint definition. Other works focus mainly on compression quality and do not provide recovered real throughput evidence. This supports the motivation in Sec.~\ref{sec:motivation} and the serving question in Sec.~\ref{sec:serving_quests}: prior speed claims are difficult to compare directly because speed evidence is not consistently reported under matched end-to-end serving conditions.

\setlength{\tabcolsep}{4pt}
\renewcommand{\arraystretch}{1.10}

\begin{table*}[t]
\centering
\footnotesize
\caption{Summary of representative SVD-style LLM compression papers (Part I).}
\label{tab:svd_compact_maintext_a}
\begin{tabularx}{\textwidth}{>{\RaggedRight\arraybackslash}p{2.25cm}>{\RaggedRight\arraybackslash}X>{\RaggedRight\arraybackslash}X>{\RaggedRight\arraybackslash}X}
\toprule
\textbf{Paper} & \textbf{Core idea} & \textbf{Speed evidence} & \textbf{How we classify it} \\
\midrule
SAES-SVD & Jointly optimizes local reconstruction and accumulated cross-layer error compensation (CEALC + ACES). & Single A6000 speedup plot on LLaMA-3-8B; workload shape under-specified. & Real hardware speedup exists, but protocol is too incomplete for strong tokens/s comparison. \\
SVD-LLM & Truncation-aware whitening + sequential low-rank weight update. & A100 + EPYC throughput sweeps with explicit batch/seq settings; repo eval uses \texttt{use\_cache=True}. & One of the cleanest single-device autoregressive benchmarks; still not a serving study. \\
ASVD & Activation-aware transformation handles outliers; iterative calibration; extends to KV-cache compression. & No real throughput benchmark recovered. & Method/quality paper, not a speed paper. \\
Dobi-SVD & Differentiable rank/truncation selection + optimal weight update + optional remapping. & A100 throughput plus TITAN Xp 12GB case study. & Real speed paper, but TITAN Xp result is offload-confounded; remapping alters ratio semantics. \\
Basis Sharing & Shared basis vectors across layers with layer-specific coefficients. & Single A100, batch $=512$, seq $=32$, one-token generation. & Decode-side microbenchmark at one operating point. \\
DipSVD & Dual-level importance protection (local singular-vector protection + global layer allocation). & Real A100-class throughput figure with batch/seq sweeps. & Real throughput paper, but protocol is less explicit than SVD-LLM. \\
AdaSVD & Adaptive error compensation (adaComp) + adaptive layer-wise ratios (adaCR). & No dedicated speed benchmark recovered. & Method paper; not directly comparable on throughput. \\
AA-SVD / AA-SVDq & Anchored compression using both original outputs and shifted inputs; block-wise end-to-end refinement; reports both no-remapping and remapped variants. & No real throughput benchmark recovered. & ``End-to-end'' here means block-level compression/refinement, not serving latency. \\
\bottomrule
\end{tabularx}
\end{table*}

\begin{table*}[t]
\centering
\footnotesize
\caption{Summary of representative SVD-style LLM compression papers (Part II).}
\label{tab:svd_compact_maintext_b}
\begin{tabularx}{\textwidth}{>{\RaggedRight\arraybackslash}p{2.25cm}>{\RaggedRight\arraybackslash}X>{\RaggedRight\arraybackslash}X>{\RaggedRight\arraybackslash}X}
\toprule
\textbf{Paper} & \textbf{Core idea} & \textbf{Speed evidence} & \textbf{How we classify it} \\
\midrule
Swift-SVD & Closed-form activation-aware compression with one eigendecomposition; dynamic rank allocation. & Single 5090 inference benchmark + end-to-end compression-time benchmark. & Strong efficiency paper, but still a microbenchmark rather than serving evaluation. \\
OBD-LLM & Hessian-aware bi-directional whitening using input and output information. & Decomposition-runtime latency only. & Not an inference-throughput paper. \\
D-Rank & Layer-wise dynamic rank allocation under a fixed global budget. & Real throughput figure exists, but protocol is under-specified in accessible text. & Useful method paper; weaker system reproducibility. \\
DF-SVD & Fast rank selection / recovery dynamics tied to singular-spectrum structure and Hessian isotropy. & Reports both compression-runtime and inference throughput. & Stronger-than-average efficiency reporting; still not serving-style evaluation. \\
CPSVD & Preserves hard columns and factorizes only easy columns; supports within-layer non-uniform compression. & Single RTX 3090, batch $=4$, seq $=1024$. & Real hardware at one operating point only. \\
ERC-SVD (ResSVD) & Residual-compensated SVD with selective late-layer compression. & A100 throughput figure, seq $=32$, varying batch size. & Real throughput paper; protocol not fully explicit. \\
MoDeGPT & Module-level decomposition of matrix pairs within transformer blocks (Nystr\"om / CR / SVD). & Single A100 80GB; batch $=256$, seq $=256$; appendix explicitly says KVCache is used. & Strong single-device generation benchmark with explicit KV-cache usage. \\
GFWSVD & Fisher-aware generalized SVD using scalable Kronecker-factored full-Fisher approximation. & A100 80GB throughput + per-token latency tables (appendix-heavy). & Real hardware speed paper, but less immediately reproducible than the clearest baselines. \\
\bottomrule
\end{tabularx}
\end{table*}

\subsection{Cross-Device Serving Robustness}

\paragraph{Cross-device serving robustness.}
Tables~\ref{tab:serving_a100_keep08}--\ref{tab:serving_a5000_keep08} report the serving summary across different GPU devices under matched request profiles. The results should not be read as a separate leaderboard because method coverage differs by device and some artifacts are not loadable under the shared vLLM path. The main trend is consistent with Sec.~\ref{sec:serving_quests}: low-rank checkpoints can improve some prefill-heavy measurements, but end-to-end latency and generation throughput remain sensitive to workload, hardware, and execution path. These tables therefore support the claim that nominal low-rank savings do not by themselves guarantee stable serving speedups.

\begin{table*}[t]
    \centering
    \caption{\textbf{A100 end-to-end serving summary for the Llama-3.1-8B keep-ratio-0.8 \texttt{bf16} benchmark.}
    Lower TTFT/E2E is better; higher prompt and generation throughput is better.}
    \label{tab:serving_summary}\label{tab:serving_a100_keep08}
    \scriptsize
    \renewcommand{\arraystretch}{1.08}
    \setlength{\tabcolsep}{3.8pt}
    \resizebox{\textwidth}{!}{
    \begin{tabular}{lcccc|cccc|cccc}
        \toprule
        & \multicolumn{4}{c|}{Prefill-heavy ($4096 \rightarrow 32$)}
        & \multicolumn{4}{c|}{Balanced ($2048 \rightarrow 128$)}
        & \multicolumn{4}{c}{Decode-heavy ($512 \rightarrow 512$)} \\
        \cmidrule(lr){2-5} \cmidrule(lr){6-9} \cmidrule(lr){10-13}
        Method
        & TTFT$\downarrow$ & E2E$\downarrow$ & In tok/s$\uparrow$ & Out tok/s$\uparrow$
        & TTFT$\downarrow$ & E2E$\downarrow$ & In tok/s$\uparrow$ & Out tok/s$\uparrow$
        & TTFT$\downarrow$ & E2E$\downarrow$ & In tok/s$\uparrow$ & Out tok/s$\uparrow$ \\
        \midrule
        Dense BF16         & 1033.71 & \textbf{3507.69} & \textbf{11521.19} & \textbf{90.03} & \textbf{73.00} & \textbf{1730.49} & \textbf{10896.70} & \textbf{681.38} & \textbf{51.19} & \textbf{6101.17} & \textbf{769.37} & \textbf{770.88} \\
        \midrule
        ASVD               & \textbf{906.88} & 3602.72 & 11154.51 & 87.17 & 93.72 & 2981.86 & 6328.40 & 395.72 & 72.32 & 11053.07 & 423.96 & 424.79 \\
        \textsc{SVD-LLM} v1 & 976.19 & 3852.98 & 10439.98 & 81.58 & 94.42 & 3031.04 & 6232.10 & 389.70 & 73.24 & 11251.25 & 416.68 & 417.49 \\
        \textsc{SVD-LLM} v2 & 967.90 & 3831.87 & 10490.16 & 81.97 & 101.52 & 3101.10 & 6058.19 & 378.82 & 75.84 & 11557.80 & 405.04 & 405.83 \\
        DoBi-SVD           & 1011.50 & 3943.80 & 10222.12 & 79.88 & 94.96 & 2712.51 & 6957.39 & 435.05 & 69.85 & 10040.88 & 467.48 & 468.39 \\
        Basis Sharing      & 1092.77 & 4295.30 & 9372.91 & 73.24 & 111.55 & 3278.81 & 5747.93 & 359.42 & 79.05 & 12253.39 & 383.41 & 384.16 \\
        \midrule
        LLM-Pruner & 368.71 & 730.42 & 6643.60 & 51.92 & 24.63 & 1506.62 & 1358.97 & 84.98 & 21.98 & 6012.41 & 84.99 & 85.15 \\
        SliceGPT   & 320.05 & 729.26 & 6477.84 & 50.62 & 24.52 & 1692.76 & 1209.10 & 75.61 & 22.08 & 6763.40 & 75.57 & 75.72 \\
        \bottomrule
    \end{tabular}}

    \vspace{3pt}
    \parbox{0.97\textwidth}{\footnotesize
    \textbf{Notes.}
    BlockPruner is omitted because the current generic vLLM loader fails on its wrapped \texttt{lm\_head} layout, although a non-vLLM Transformers manual-decode fallback completes.}
\end{table*}

\begin{table*}[t]
    \centering
    \caption{\textbf{L40S end-to-end serving summary for the Llama-3.1-8B keep-ratio-0.8 \texttt{bf16} benchmark.}
    Lower TTFT/E2E is better; higher prompt and generation throughput is better.}
    \label{tab:serving_l40s_keep08}
    \scriptsize
    \renewcommand{\arraystretch}{1.08}
    \setlength{\tabcolsep}{3.8pt}
    \resizebox{\textwidth}{!}{
    \begin{tabular}{lcccc|cccc|cccc}
        \toprule
        & \multicolumn{4}{c|}{Prefill-heavy ($4096 \rightarrow 32$)}
        & \multicolumn{4}{c|}{Balanced ($2048 \rightarrow 128$)}
        & \multicolumn{4}{c}{Decode-heavy ($512 \rightarrow 512$)} \\
        \cmidrule(lr){2-5} \cmidrule(lr){6-9} \cmidrule(lr){10-13}
        Method
        & TTFT$\downarrow$ & E2E$\downarrow$ & In tok/s$\uparrow$ & Out tok/s$\uparrow$
        & TTFT$\downarrow$ & E2E$\downarrow$ & In tok/s$\uparrow$ & Out tok/s$\uparrow$
        & TTFT$\downarrow$ & E2E$\downarrow$ & In tok/s$\uparrow$ & Out tok/s$\uparrow$ \\
        \midrule
        Dense BF16         & \textbf{932.72} & \textbf{3793.12} & \textbf{10583.97} & \textbf{82.71} & 433.55 & 4179.39 & 4592.98 & 287.20 & 85.81 & 11919.05 & 393.47 & 394.24 \\
        \midrule
        ASVD               & 1046.45 & 4248.33 & 9467.95 & 73.99 & 406.70 & 4612.59 & 4188.58 & 261.91 & \textbf{75.98} & 14164.50 & 335.53 & 336.19 \\
        \textsc{SVD-LLM} v1 & 1027.94 & 4111.04 & 9788.40 & 76.49 & 392.13 & 3883.31 & 4940.85 & 308.95 & 82.92 & 11251.31 & 417.29 & 418.11 \\
        \textsc{SVD-LLM} v2 & 1093.06 & 4436.42 & 9060.56 & 70.80 & 434.85 & 4859.52 & 3966.78 & 248.04 & 76.79 & 14803.03 & 320.04 & 320.66 \\
        DoBi-SVD           & 1001.48 & 4006.10 & 10045.33 & 78.50 & \textbf{384.81} & \textbf{3781.46} & \textbf{5074.26} & \textbf{317.30} & 77.27 & \textbf{10897.85} & \textbf{430.63} & \textbf{431.48} \\
        Basis Sharing      & 1230.96 & 4951.35 & 8137.69 & 63.59 & 544.16 & 5474.86 & 3550.74 & 222.03 & 86.49 & 16575.79 & 288.80 & 289.36 \\
        \midrule
        LLM-Pruner$^\dagger$ & 315.92 & 887.53 & 5152.54 & 40.26 & 30.03 & 2374.65 & 861.98 & 53.90 & 27.04 & 9465.31 & 53.99 & 54.09 \\
        SliceGPT$^\dagger$   & 316.84 & 940.78 & 4833.34 & 37.77 & 31.86 & 2592.38 & 789.55 & 49.37 & 28.57 & 10336.09 & 49.44 & 49.53 \\
        \bottomrule
    \end{tabular}}

\end{table*}

\begin{table*}[t]
    \centering
    \caption{\textbf{RTX A5000 end-to-end serving.}
    Lower TTFT/E2E is better; higher prompt and generation throughput is better.}
    \label{tab:serving_a5000_keep08}
    \scriptsize
    \renewcommand{\arraystretch}{1.08}
    \setlength{\tabcolsep}{3.8pt}
    \resizebox{\textwidth}{!}{
    \begin{tabular}{lcccc|cccc|cccc}
        \toprule
        & \multicolumn{4}{c|}{Prefill-heavy ($4096 \rightarrow 32$)}
        & \multicolumn{4}{c|}{Balanced ($2048 \rightarrow 128$)}
        & \multicolumn{4}{c}{Decode-heavy ($512 \rightarrow 512$)} \\
        \cmidrule(lr){2-5} \cmidrule(lr){6-9} \cmidrule(lr){10-13}
        Method
        & TTFT$\downarrow$ & E2E$\downarrow$ & In tok/s$\uparrow$ & Out tok/s$\uparrow$
        & TTFT$\downarrow$ & E2E$\downarrow$ & In tok/s$\uparrow$ & Out tok/s$\uparrow$
        & TTFT$\downarrow$ & E2E$\downarrow$ & In tok/s$\uparrow$ & Out tok/s$\uparrow$ \\
        \midrule
        Dense BF16             & 2426.34 & \textbf{8668.29} & \textbf{4722.81} & \textbf{36.90} & \textbf{1509.29} & \textbf{7298.74} & \textbf{2717.86} & \textbf{169.87} & \textbf{623.80} & \textbf{14447.66} & \textbf{328.66} & \textbf{328.66} \\
        \midrule
        ASVD                   & 2544.26 & 10977.00 & 3739.81 & 29.22 & 1934.54 & 9760.96 & 2025.96 & 126.62 & 957.00 & 20533.82 & 230.95 & 230.95 \\
        \textsc{SVD-LLM} v1    & 2231.26 & 9636.49 & 4254.03 & 33.23 & 1677.82 & 8998.35 & 2180.49 & 136.28 & 797.32 & 20901.97 & 224.94 & 224.94 \\
        \textsc{SVD-LLM} v2    & 2639.96 & 11344.07 & 3625.66 & 28.33 & 2000.00 & 9878.82 & 1996.81 & 124.80 & 980.79 & 21198.33 & 222.23 & 222.23 \\
        DoBi-SVD               & \textbf{2123.16} & 9163.42 & 4470.79 & 34.93 & 1593.67 & 8790.49 & 2239.36 & 139.96 & 725.63 & 20962.44 & 224.96 & 224.96 \\
        Basis Sharing          & 2839.16 & 12318.84 & 3352.18 & 26.19 & 2177.91 & 11603.11 & 1725.16 & 107.82 & 1062.42 & 26087.88 & 186.01 & 186.01 \\
        \midrule
        SliceGPT               & 1278.76 & 2419.43 & 1995.06 & 15.59 & 100.80 & 5030.57 & 190.75 & 11.92 & 91.92 & 20075.97 & 25.55 & 25.55 \\
        LLM-Pruner             & 1092.81 & 1859.03 & 2662.71 & 20.80 & 77.49 & 3285.21 & 621.69 & 38.86 & 65.53 & 13164.42 & 38.67 & 38.67 \\
        \bottomrule
    \end{tabular}}
\end{table*}

\subsection{Large-Model Fixed-Budget Feasibility}

\paragraph{Large-model compression feasibility.}
Table~\ref{tab:a1_large_model_feasibility} reports released-code artifact-generation readiness for Llama-3.1-70B under a fixed one-H200 recipe. The audited methods all have working 7B/8B artifacts in the \benchname{} evaluation path, but the released 70B source-code paths do not directly produce standardized \benchname{} artifacts out of the box. The blockers are implementation-level rather than evidence that the underlying algorithms cannot scale: ASVD is stopped by a legacy \texttt{lm\_eval} API dependency after loading the model, \textsc{SVD-LLM} v1 and DoBi-SVD run into H200-scale memory pressure during whitening or training, Basis Sharing fail in the CUDA eigensolver path, and MoDeGPT fails when an Accelerate-offloaded model is moved back to GPU. These failures show that the released codebases are not yet robust artifact-generation pipelines at 70B under the stated recipe. They should therefore be interpreted as source-code readiness and fixed-budget feasibility results, not as method-quality claims or proofs of algorithmic impossibility with dependency fixes, numerical fallbacks, custom offload, sharding, or additional engineering.

\begin{table*}[t]
\centering
\caption{\textbf{Released-code artifact-generation readiness for 70B on one H200.}
Rows audit whether the available implementation can directly produce a standardized
\textsc{LowRankArena} artifact for Llama-3.1-70B at keep ratio 0.6, using one
H200-141GB accelerator and no multi-GPU sharding or method-specific repair.}
\label{tab:a1_large_model_feasibility}
\footnotesize
\renewcommand{\arraystretch}{1.13}
\setlength{\tabcolsep}{3.0pt}
\begin{tabular*}{\textwidth}{@{\extracolsep{\fill}} l >{\raggedright\arraybackslash}p{2.25cm} c c >{\raggedright\arraybackslash}p{5.9cm} @{}}
\toprule
Method & Source-model scope & 7B/8B & Peak VRAM & 70B one-H200 evidence \\
\midrule
ASVD
& Hugging Face \texttt{AutoModel}; Llama-style scripts
& Ready
& 74.2 GiB
& Loads Llama-3.1-70B under the H200 memory cap, but stops on a legacy
\texttt{lm\_eval.base} import inside the sensitivity path before artifact export. \\

\textsc{SVD-LLM} v1
& LLaMA/OPT components
& Ready
& 139.1 GiB
& Enters the low-resource whitening path, then OOMs during
whitening/eigendecomposition near the H200 memory ceiling; no artifact is produced. \\


Basis Sharing
& LLaMA/OPT/Mistral configs
& Ready
& 15.8 GiB
& Enters the local low-resource profiling path, then fails in the same CUDA
eigensolver path as \textsc{SVD-LLM} v2 before artifact export. \\

MoDeGPT
& Llama/OPT/Qwen adapters
& Ready
& 123.6 GiB
& Loads Llama-3.1-70B, then fails when moving an Accelerate-offloaded model to
GPU before compression; no artifact is produced. \\

DoBi-SVD
& \texttt{AutoModel} updater for decoder LMs
& Ready
& 139.1 GiB
& The outer wrapper exits successfully, but the internal trainer reaches the H200
memory ceiling and records trainer OOM before training/export, so no artifact is
counted. \\
\bottomrule
\end{tabular*}

\vspace{2pt}
\parbox{0.97\textwidth}{\footnotesize
\textbf{Notes.}
``Ready'' means local 7B/8B artifact generation completed and the resulting
checkpoint can enter the normal \textsc{LowRankArena} evaluation path.
The 70B column reports released-code readiness under the stated one-H200 recipe.
Blocked rows should not be read as algorithmic impossibility with dependency
fixes, numerical fallbacks, custom offload, or multi-GPU sharding.}
\end{table*}

\subsection{Expanded Base-Model Leaderboards}

\paragraph{Llama-2-7B standardized base leaderboard.}
Table~\ref{tab:llama2_leaderboard} extends the standardized base-model comparison to the full keep-ratio grid on Llama-2-7B. The results show that method ordering is not fixed across metrics or budgets. MoDeGPT is strong at higher keep ratios and remains strong on MCQ Avg., but the best perplexity profile changes as compression becomes more aggressive. The table also shows that answer-selection accuracy can remain relatively stable while WikiText-2 and C4 perplexity degrade.

\begin{table*}[t]
    \centering
    \caption{\textbf{Llama-2-7B leaderboard.}
All methods are evaluated from normalized summaries under the primary full-precision comparison scope with matched task versions, evaluation scripts, and budget definition.
The table reports the full 80/70/60/50/40 keep-ratio grid with perplexity, MCQ Avg., and base-math metrics.}
    \scriptsize
    \renewcommand{\arraystretch}{1.10}
    \setlength{\tabcolsep}{2pt}
    \label{tab:llama2_leaderboard}
    \scriptsize
    \renewcommand{\arraystretch}{1.10}
    \setlength{\tabcolsep}{3pt}
    \resizebox{\textwidth}{!}{
    \begin{tabular}{@{}lll!{\vrule width 0.45pt}cc!{\vrule width 0.45pt}ccccccc!{\vrule width 0.45pt}c!{\vrule width 0.45pt}cc@{}}
        \toprule
        & & &
        \multicolumn{2}{c!{\vrule width 0.45pt}}{\textbf{Perplexity} $\downarrow$} &
        \multicolumn{7}{c!{\vrule width 0.45pt}}{\textbf{General MCQ} $\uparrow$} &
        \multicolumn{1}{c!{\vrule width 0.45pt}}{\textbf{MCQ Avg.} $\uparrow$} &
        \multicolumn{2}{c@{}}{\textbf{Math} $\uparrow$} \\
        \cmidrule(lr){4-5}\cmidrule(lr){6-12}\cmidrule(lr){13-13}\cmidrule(l){14-15}
        Model & Budget & Method
        & WikiT. & C4
        & BoolQ & ARC-E & ARC-C & WinoG. & PIQA & HellaS. & OBQA
        & MCQ Avg.
        & MathQA & MMLU-M \\
        \midrule

        \rowcolor{gray!10}
        \multirow{31}{*}{Llama-2-7B}
            & \multicolumn{2}{l}{\textbf{Dense FP}}
            & 5.47 & 7.11
            & 0.794 & 0.738 & 0.450 & 0.694 & 0.787 & 0.762 & 0.440
            & 0.666
            & 0.283 & 0.288 \\
        \cmidrule(lr){2-15}

            & \multirow{6}{*}{80\%} & ASVD
            & 10.82 & 15.04
            & \textbf{0.685} & 0.562 & 0.333 & 0.639 & 0.709 & 0.620 & 0.382
            & 0.561
            & 0.248 & \textbf{0.304} \\
            & & \textsc{SVD-LLM} v1
            & 8.39 & 20.20
            & 0.430 & 0.490 & 0.295 & 0.622 & 0.671 & 0.550 & 0.356
            & 0.488
            & 0.234 & 0.251 \\
            & & \textsc{SVD-LLM} v2
            & 9.77 & 24.76
            & 0.485 & 0.421 & 0.282 & 0.606 & 0.631 & 0.505 & 0.328
            & 0.465
            & 0.227 & 0.297 \\
            & & DoBi-SVD
            & 10.06 & 25.27
            & 0.463 & 0.470 & 0.283 & 0.618 & 0.645 & 0.477 & 0.336
            & 0.470
            & 0.228 & 0.259 \\
            & & Basis Sharing
            & 7.71 & 16.42
            & 0.565 & 0.571 & 0.337 & 0.644 & 0.708 & 0.601 & 0.390
            & 0.545
            & 0.253 & 0.266 \\
            & & MoDeGPT
            & \textbf{6.88} & \textbf{10.97}
            & 0.613 & \textbf{0.659} & \textbf{0.416} & \textbf{0.691} & \textbf{0.762} & \textbf{0.702} & \textbf{0.402}
            & \textbf{0.606}
            & \textbf{0.271} & 0.288 \\

        \cmidrule(lr){2-15}

            & \multirow{6}{*}{70\%} & ASVD
            & 468.47 & 669.87
            & 0.463 & 0.282 & 0.253 & 0.487 & 0.526 & 0.291 & 0.256
            & 0.366
            & 0.200 & 0.260 \\
            & & \textsc{SVD-LLM} v1
            & 10.67 & 34.53
            & 0.421 & 0.442 & 0.275 & 0.580 & 0.621 & 0.462 & 0.346
            & 0.450
            & 0.222 & 0.295 \\
            & & \textsc{SVD-LLM} v2
            & 14.79 & 55.50
            & 0.404 & 0.356 & 0.251 & 0.572 & 0.564 & 0.376 & 0.318
            & 0.406
            & 0.223 & \textbf{0.296} \\
            & & DoBi-SVD
            & 12.56 & 35.14
            & 0.467 & 0.410 & 0.259 & 0.581 & 0.616 & 0.416 & 0.328
            & 0.439
            & 0.229 & 0.261 \\
            & & Basis Sharing
            & 9.70 & 25.65
            & 0.434 & 0.520 & 0.288 & 0.620 & 0.661 & 0.505 & 0.360
            & 0.484
            & 0.241 & 0.256 \\
            & & MoDeGPT
            & \textbf{8.58} & \textbf{14.83}
            & \textbf{0.635} & \textbf{0.572} & \textbf{0.363} & \textbf{0.666} & \textbf{0.705} & \textbf{0.645} & \textbf{0.384}
            & \textbf{0.567}
            & \textbf{0.263} & 0.274 \\

        \cmidrule(lr){2-15}

            & \multirow{6}{*}{60\%} & ASVD
            & 4517.97 & 4339.12
            & 0.389 & 0.269 & 0.271 & 0.499 & 0.493 & 0.255 & 0.254
            & 0.347
            & 0.204 & 0.202 \\
            & & \textsc{SVD-LLM} v1
            & 570817.53 & 586576.93
            & 0.391 & 0.368 & 0.243 & 0.557 & 0.558 & 0.372 & 0.320
            & 0.401
            & 0.220 & 0.293 \\
            & & \textsc{SVD-LLM} v2
            & 26.71 & 133.12
            & 0.378 & 0.309 & 0.232 & 0.506 & 0.528 & 0.318 & 0.288
            & 0.366
            & 0.219 & \textbf{0.299} \\
            & & DoBi-SVD
            & 17.50 & 61.49
            & 0.389 & 0.353 & 0.252 & 0.557 & 0.566 & 0.358 & 0.278
            & 0.393
            & 0.228 & 0.296 \\
            & & Basis Sharing
            & \textbf{13.88} & \textbf{46.35}
            & 0.392 & 0.443 & 0.259 & 0.594 & 0.612 & 0.416 & 0.340
            & 0.436
            & 0.232 & 0.252 \\
            & & MoDeGPT
            & 27.27 & 93.72
            & \textbf{0.622} & \textbf{0.474} & \textbf{0.318} & \textbf{0.645} & \textbf{0.653} & \textbf{0.540} & \textbf{0.342}
            & \textbf{0.514}
            & \textbf{0.242} & 0.266 \\

        \cmidrule(lr){2-15}

            & \multirow{6}{*}{50\%} & ASVD
            & 24894.76 & 24394.20
            & 0.424 & 0.261 & 0.270 & 0.491 & 0.499 & 0.260 & 0.258
            & 0.352
            & 0.210 & 0.237 \\
            & & \textsc{SVD-LLM} v1
            & 182643.36 & 230247.50
            & 0.378 & 0.310 & 0.232 & 0.541 & 0.528 & 0.311 & 0.270
            & 0.367
            & 0.211 & 0.285 \\
            & & \textsc{SVD-LLM} v2
            & 55.57 & 282.01
            & 0.378 & 0.286 & 0.247 & 0.500 & 0.523 & 0.293 & 0.246
            & 0.353
            & 0.211 & 0.276 \\
            & & DoBi-SVD
            & 25.84 & \textbf{88.99}
            & 0.386 & 0.301 & 0.240 & 0.534 & 0.548 & 0.326 & 0.282
            & 0.374
            & 0.216 & \textbf{0.296} \\
            & & Basis Sharing
            & \textbf{23.70} & 109.43
            & 0.378 & \textbf{0.364} & 0.244 & 0.556 & 0.548 & 0.339 & \textbf{0.294}
            & 0.389
            & \textbf{0.223} & 0.283 \\
            & & MoDeGPT
            & 73.01 & 330.81
            & \textbf{0.606} & 0.342 & \textbf{0.299} & \textbf{0.593} & \textbf{0.586} & \textbf{0.423} & 0.286
            & \textbf{0.448}
            & 0.202 & 0.218 \\

        \cmidrule(lr){2-15}

            & \multirow{6}{*}{40\%} & ASVD
            & 40884.78 & 50316.43
            & 0.385 & 0.266 & \textbf{0.286} & 0.507 & 0.496 & 0.265 & 0.250
            & 0.351
            & 0.194 & 0.273 \\
            & & \textsc{SVD-LLM} v1
            & 330368.91 & 301017.68
            & 0.378 & 0.274 & 0.256 & 0.493 & 0.517 & 0.285 & 0.238
            & 0.349
            & 0.204 & 0.260 \\
            & & \textsc{SVD-LLM} v2
            & 96.07 & 475.84
            & 0.378 & 0.276 & 0.244 & 0.493 & 0.522 & 0.282 & 0.246
            & 0.349
            & 0.208 & 0.235 \\
            & & DoBi-SVD
            & \textbf{43.06} & \textbf{146.22}
            & 0.378 & 0.274 & 0.258 & 0.513 & 0.517 & 0.295 & 0.266
            & 0.357
            & 0.210 & \textbf{0.303} \\
            & & Basis Sharing
            & 50.33 & 270.93
            & 0.378 & \textbf{0.295} & 0.231 & 0.520 & 0.519 & 0.293 & 0.260
            & 0.357
            & \textbf{0.213} & 0.284 \\
            & & MoDeGPT
            & 116.39 & 663.03
            & \textbf{0.458} & 0.287 & 0.279 & \textbf{0.567} & \textbf{0.533} & \textbf{0.336} & \textbf{0.282}
            & \textbf{0.392}
            & 0.211 & 0.290 \\

        \bottomrule
    \end{tabular}}

\end{table*}

\paragraph{Qwen3-8B-Base standardized leaderboard.}
Table~\ref{tab:qwen3_8b_base_detailed_leaderboard} shows that the Llama-2 trend does not transfer unchanged to Qwen3-8B-Base. ASVD is strong at the highest keep ratio, while Basis Sharing, MoDeGPT, and SVD-LLM variants become competitive in different metric groups at lower keep ratios. Several methods also show large WikiText-2 and C4 degradation even when MCQ Avg. remains comparable. This supports the main claim that broader models and matched benchmarks are needed to clarify conditional rankings and avoid drawing a single-method progress claim from one model or one metric.

\begin{table*}[t]
    \centering
    \caption{\textbf{Qwen3-8B-Base leaderboard.}
All methods are evaluated from normalized summaries under the primary full-precision comparison scope with matched task versions, evaluation scripts, and budget definition.
The table reports the full 80/70/60/50/40 keep-ratio grid with perplexity, MCQ Avg., and base-math metrics.}
    \label{tab:qwen3_8b_base_detailed_leaderboard}
    \scriptsize
    \renewcommand{\arraystretch}{1.10}
    \setlength{\tabcolsep}{3pt}
    \resizebox{\textwidth}{!}{
    \begin{tabular}{@{}lll!{\vrule width 0.45pt}cc!{\vrule width 0.45pt}ccccccc!{\vrule width 0.45pt}c!{\vrule width 0.45pt}cc@{}}
        \toprule
        & & &
        \multicolumn{2}{c!{\vrule width 0.45pt}}{\textbf{Perplexity} $\downarrow$} &
        \multicolumn{7}{c!{\vrule width 0.45pt}}{\textbf{General MCQ} $\uparrow$} &
        \multicolumn{1}{c!{\vrule width 0.45pt}}{\textbf{MCQ Avg.} $\uparrow$} &
        \multicolumn{2}{c@{}}{\textbf{Math} $\uparrow$} \\
        \cmidrule(lr){4-5}\cmidrule(lr){6-12}\cmidrule(lr){13-13}\cmidrule(l){14-15}
        Model & Budget & Method
        & WikiT. & C4
        & BoolQ & ARC-E & ARC-C & WinoG. & PIQA & HellaS. & OBQA
        & MCQ Avg.
        & MathQA & MMLU-M \\
        \midrule

        \rowcolor{gray!10}
        \multirow{31}{*}{Qwen3-8B-Base}
            & \multicolumn{2}{l}{\textbf{Dense FP}}
            & 7.00 & 11.78
            & 0.830 & 0.800 & 0.570 & 0.727 & 0.793 & 0.787 & 0.420
            & 0.704
            & 0.542 & 0.729 \\
        \cmidrule(lr){2-15}

            & \multirow{6}{*}{80\%} & ASVD
            & 11.88 & \textbf{20.54}
            & \textbf{0.792} & \textbf{0.758} & \textbf{0.483} & 0.651 & \textbf{0.745} & 0.641 & \textbf{0.414}
            & \textbf{0.641}
            & \textbf{0.420} & \textbf{0.460} \\
            & & \textsc{SVD-LLM} v1
            & 11.08 & 33.90
            & 0.688 & 0.649 & 0.424 & 0.669 & 0.712 & 0.616 & 0.408
            & 0.595
            & 0.330 & 0.355 \\
            & & \textsc{SVD-LLM} v2
            & 11.59 & 35.63
            & 0.669 & 0.618 & 0.411 & 0.654 & 0.715 & 0.609 & 0.398
            & 0.582
            & 0.321 & 0.352 \\
            & & DoBi-SVD
            & 51.21 & 236.65
            & 0.460 & 0.335 & 0.284 & 0.519 & 0.550 & 0.388 & 0.262
            & 0.400
            & 0.206 & 0.264 \\
            & & Basis Sharing
            & 11.05 & 31.67
            & 0.676 & 0.578 & 0.439 & \textbf{0.673} & 0.726 & 0.635 & 0.388
            & 0.588
            & 0.331 & 0.347 \\
            & & MoDeGPT
            & \textbf{10.34} & 22.46
            & 0.658 & 0.613 & 0.430 & 0.655 & 0.731 & \textbf{0.683} & 0.408
            & 0.597
            & 0.279 & 0.297 \\

        \cmidrule(lr){2-15}

            & \multirow{6}{*}{70\%} & ASVD
            & 99.51 & 147.69
            & 0.622 & 0.448 & 0.276 & 0.519 & 0.607 & 0.356 & 0.302
            & 0.447
            & 0.244 & 0.271 \\
            & & \textsc{SVD-LLM} v1
            & 13.82 & 55.62
            & 0.624 & \textbf{0.527} & 0.340 & 0.609 & 0.663 & 0.499 & 0.358
            & 0.517
            & 0.268 & 0.276 \\
            & & \textsc{SVD-LLM} v2
            & 15.74 & 70.05
            & 0.621 & 0.496 & 0.336 & 0.597 & 0.649 & 0.479 & 0.346
            & 0.503
            & 0.271 & \textbf{0.304} \\
            & & DoBi-SVD
            & 867.24 & 908.97
            & 0.420 & 0.276 & 0.239 & 0.523 & 0.535 & 0.290 & 0.282
            & 0.366
            & 0.208 & 0.278 \\
            & & Basis Sharing
            & 13.38 & 51.15
            & \textbf{0.627} & 0.522 & \textbf{0.365} & \textbf{0.642} & 0.666 & 0.535 & \textbf{0.372}
            & \textbf{0.533}
            & \textbf{0.277} & 0.297 \\
            & & MoDeGPT
            & \textbf{12.81} & \textbf{32.74}
            & 0.570 & 0.485 & 0.337 & 0.613 & \textbf{0.670} & \textbf{0.576} & 0.362
            & 0.516
            & 0.238 & 0.284 \\

        \cmidrule(lr){2-15}

            & \multirow{6}{*}{60\%} & ASVD
            & 1359.38 & 1484.57
            & 0.502 & 0.292 & 0.241 & 0.499 & 0.544 & 0.281 & 0.270
            & 0.375
            & 0.212 & 0.234 \\
            & & \textsc{SVD-LLM} v1
            & 20.44 & 112.01
            & 0.544 & 0.382 & 0.265 & 0.545 & 0.590 & 0.385 & 0.266
            & 0.425
            & 0.231 & \textbf{0.307} \\
            & & \textsc{SVD-LLM} v2
            & 29.25 & 188.58
            & 0.593 & 0.334 & 0.245 & 0.545 & 0.581 & 0.359 & 0.292
            & 0.421
            & 0.228 & 0.295 \\
            & & DoBi-SVD
            & 107.63 & 610.21
            & 0.511 & 0.290 & 0.253 & 0.511 & 0.523 & 0.310 & 0.306
            & 0.386
            & 0.203 & 0.302 \\
            & & Basis Sharing
            & 18.59 & 95.39
            & \textbf{0.630} & \textbf{0.457} & 0.278 & \textbf{0.580} & 0.620 & 0.423 & 0.298
            & \textbf{0.469}
            & \textbf{0.253} & 0.302 \\
            & & MoDeGPT
            & \textbf{18.40} & \textbf{55.66}
            & 0.509 & 0.414 & \textbf{0.293} & 0.569 & \textbf{0.625} & \textbf{0.460} & \textbf{0.328}
            & 0.457
            & 0.228 & 0.246 \\

        \cmidrule(lr){2-15}

            & \multirow{6}{*}{50\%} & ASVD
            & 8442.35 & 9784.64
            & 0.470 & 0.265 & \textbf{0.262} & 0.518 & 0.508 & 0.263 & 0.288
            & 0.368
            & 0.198 & 0.224 \\
            & & \textsc{SVD-LLM} v1
            & 35.31 & 222.81
            & 0.392 & 0.312 & 0.228 & \textbf{0.548} & 0.557 & 0.322 & 0.256
            & 0.374
            & 0.233 & 0.301 \\
            & & \textsc{SVD-LLM} v2
            & 68.99 & 467.68
            & 0.424 & 0.295 & 0.225 & 0.519 & 0.534 & 0.301 & 0.260
            & 0.365
            & \textbf{0.236} & \textbf{0.305} \\
            & & DoBi-SVD
            & 2467.24 & 2008.83
            & 0.495 & 0.276 & 0.254 & 0.504 & 0.511 & 0.276 & 0.290
            & 0.372
            & 0.206 & 0.195 \\
            & & Basis Sharing
            & \textbf{33.81} & 249.67
            & \textbf{0.499} & 0.340 & 0.241 & 0.539 & 0.563 & 0.339 & 0.286
            & \textbf{0.401}
            & 0.234 & 0.289 \\
            & & MoDeGPT
            & 36.04 & \textbf{115.98}
            & 0.381 & \textbf{0.357} & 0.254 & 0.520 & \textbf{0.573} & \textbf{0.359} & \textbf{0.298}
            & 0.392
            & 0.211 & 0.242 \\

        \cmidrule(lr){2-15}

            & \multirow{6}{*}{40\%} & ASVD
            & 32908.39 & 52680.88
            & 0.410 & 0.253 & \textbf{0.259} & 0.515 & 0.529 & 0.264 & 0.278
            & 0.358
            & 0.197 & \textbf{0.298} \\
            & & \textsc{SVD-LLM} v1
            & \textbf{73.92} & 512.54
            & 0.378 & 0.283 & 0.240 & 0.505 & 0.531 & 0.290 & 0.242
            & 0.353
            & 0.226 & 0.295 \\
            & & \textsc{SVD-LLM} v2
            & 572.04 & 2580.88
            & 0.379 & 0.272 & 0.245 & 0.491 & 0.510 & 0.272 & 0.276
            & 0.349
            & 0.210 & 0.200 \\
            & & DoBi-SVD
            & 3116.15 & 2313.29
            & \textbf{0.543} & 0.267 & 0.251 & \textbf{0.519} & 0.526 & 0.271 & 0.276
            & \textbf{0.379}
            & 0.208 & 0.231 \\
            & & Basis Sharing
            & 75.94 & 619.66
            & 0.378 & 0.301 & 0.235 & 0.516 & 0.519 & 0.300 & 0.250
            & 0.357
            & \textbf{0.228} & 0.268 \\
            & & MoDeGPT
            & 93.23 & \textbf{272.39}
            & 0.378 & \textbf{0.319} & 0.243 & 0.508 & \textbf{0.552} & \textbf{0.307} & \textbf{0.292}
            & 0.371
            & 0.216 & 0.280 \\

        \bottomrule
    \end{tabular}}

\end{table*}

\subsection{Auxiliary Instruction-Tuned Leaderboards}

\paragraph{Llama-3.1-8B-Instruct downstream leaderboard.}
Table~\ref{tab:llama31_8b_instruct_reasoning_leaderboard} reports the auxiliary instruction-tuned sweep under matched scripts. The dense checkpoint remains much stronger than the compressed checkpoints. Most compressed rows lose GSM8K accuracy early, while IFEval and MMLU-Pro show smaller but still substantial degradation. MoDeGPT has the strongest average at the higher keep ratios in the completed table, but the ordering becomes less stable under stronger compression. Because instruction-tuned scores depend on prompt templates, decoding settings, and evaluator versions, this table should be used as auxiliary evidence rather than as a primary leaderboard claim.

\begin{table*}[t]
    \centering
    \caption{\textbf{Llama-3.1-8B-Instruct auxiliary evaluation.}
Methods are evaluated across keep ratios with matched task versions and evaluation scripts.
Avg. is the macro mean over the displayed instruction-style metrics.}
    \label{tab:llama31_8b_instruct_reasoning_leaderboard}
    \scriptsize
    \renewcommand{\arraystretch}{1.10}
    \setlength{\tabcolsep}{4pt}
    \begin{tabular}{lllcccc}
        \toprule
        Model & Budget & Method & IFEval Strict$\uparrow$ & GSM8K EM$\uparrow$ & MMLU-Pro$\uparrow$ & Avg$\uparrow$ \\
        \midrule
        \multirow{31}{*}{Llama-3.1-8B-Instruct}
            & FP & Dense FP            & 0.756 & 0.767 & 0.374 & 0.632 \\
            \cmidrule(lr){2-7}
            & \multirow{6}{*}{80\%}
                & ASVD                & 0.113 & 0.000 & 0.112 & 0.075 \\
            &   & \textsc{SVD-LLM} v1 & \textbf{0.128} & 0.027 & 0.122 & 0.092 \\
            &   & \textsc{SVD-LLM} v2 & 0.109 & 0.024 & 0.109 & 0.081 \\
            &   & DoBi-SVD            & 0.111 & 0.000 & 0.105 & 0.072 \\
            &   & Basis Sharing       & 0.107 & \textbf{0.047} & 0.137 & 0.097 \\
            &   & MoDeGPT             & 0.116 & 0.000 & \textbf{0.316} & \textbf{0.144} \\
            \cmidrule(lr){2-7}
            & \multirow{6}{*}{70\%}
                & ASVD                & 0.118 & 0.000 & 0.111 & 0.076 \\
            &   & \textsc{SVD-LLM} v1 & \textbf{0.152} & 0.003 & 0.112 & 0.089 \\
            &   & \textsc{SVD-LLM} v2 & 0.140 & 0.002 & 0.115 & 0.086 \\
            &   & DoBi-SVD            & 0.120 & 0.000 & 0.102 & 0.074 \\
            &   & Basis Sharing       & 0.107 & \textbf{0.005} & 0.112 & 0.075 \\
            &   & MoDeGPT             & 0.120 & 0.000 & \textbf{0.293} & \textbf{0.138} \\
            \cmidrule(lr){2-7}
            & \multirow{6}{*}{60\%}
                & ASVD                & 0.115 & 0.000 & 0.111 & 0.075 \\
            &   & \textsc{SVD-LLM} v1 & 0.152 & 0.000 & 0.116 & 0.089 \\
            &   & \textsc{SVD-LLM} v2 & \textbf{0.153} & 0.000 & 0.112 & 0.088 \\
            &   & DoBi-SVD            & 0.129 & 0.000 & 0.106 & 0.078 \\
            &   & Basis Sharing       & 0.152 & 0.000 & 0.118 & 0.090 \\
            &   & MoDeGPT             & 0.146 & 0.000 & \textbf{0.218} & \textbf{0.121} \\
            \cmidrule(lr){2-7}
            & \multirow{6}{*}{50\%}
                & ASVD                & 0.124 & 0.000 & 0.110 & 0.078 \\
            &   & \textsc{SVD-LLM} v1 & \textbf{0.152} & 0.000 & 0.113 & 0.088 \\
            &   & \textsc{SVD-LLM} v2 & 0.144 & 0.000 & 0.112 & 0.085 \\
            &   & DoBi-SVD            & 0.128 & 0.000 & 0.110 & 0.079 \\
            &   & Basis Sharing       & \textbf{0.152} & 0.000 & 0.115 & \textbf{0.089} \\
            &   & MoDeGPT             & 0.133 & 0.000 & \textbf{0.123} & 0.085 \\
            \cmidrule(lr){2-7}
            & \multirow{6}{*}{40\%}
                & ASVD                & 0.111 & 0.000 & \textbf{0.113} & 0.075 \\
            &   & \textsc{SVD-LLM} v1 & \textbf{0.150} & 0.000 & \textbf{0.113} & \textbf{0.088} \\
            &   & \textsc{SVD-LLM} v2 & 0.144 & 0.000 & \textbf{0.113} & 0.086 \\
            &   & DoBi-SVD            & 0.144 & 0.000 & 0.109 & 0.084 \\
            &   & Basis Sharing       & 0.129 & 0.000 & \textbf{0.113} & 0.081 \\
            &   & MoDeGPT             & 0.139 & 0.000 & 0.112 & 0.083 \\
        \bottomrule
    \end{tabular}
\end{table*}

\paragraph{Qwen3-8B instruction-style downstream leaderboard.}
Table~\ref{tab:qwen3_8b_reasoning_leaderboard} reports the same auxiliary instruction-style evaluation on Qwen3-8B. The dense checkpoint remains much stronger, especially on GSM8K. At the highest keep ratio, Basis Sharing has the strongest average among compressed rows, while ASVD, SVD-LLM variants, and MoDeGPT are competitive in different metric columns or budgets. Under stronger compression, most GSM8K scores move close to floor and method differences narrow. This table should also be used as auxiliary evidence rather than as a primary leaderboard claim, because instruction-tuned scores depend on prompt templates, decoding settings, and evaluator versions.

\begin{table*}[t]
    \centering
   \caption{\textbf{Qwen3-8B instruction-style auxiliary evaluation.}
Methods are evaluated across keep ratios with matched task versions and evaluation scripts.
Avg. is the macro mean over the displayed instruction-style metrics.}
    \label{tab:qwen3_8b_reasoning_leaderboard}
    \scriptsize
    \renewcommand{\arraystretch}{1.10}
    \setlength{\tabcolsep}{4pt}
    \begin{tabular}{lllcccc}
        \toprule
        Model & Budget & Method & IFEval Strict$\uparrow$ & GSM8K EM$\uparrow$ & MMLU-Pro$\uparrow$ & Avg$\uparrow$ \\
        \midrule
        \multirow{31}{*}{Qwen3-8B}
            & FP & Dense FP            & 0.344 & 0.883 & 0.477 & 0.568 \\
            \cmidrule(lr){2-7}
            & \multirow{6}{*}{80\%}
                & ASVD                & \textbf{0.220} & 0.119 & 0.217 & 0.185 \\
            &   & \textsc{SVD-LLM} v1 & 0.115 & 0.192 & 0.246 & 0.184 \\
            &   & \textsc{SVD-LLM} v2 & 0.124 & 0.249 & 0.228 & 0.200 \\
            &   & DoBi-SVD            & 0.129 & 0.000 & 0.115 & 0.081 \\
            &   & Basis Sharing       & 0.104 & \textbf{0.318} & \textbf{0.258} & \textbf{0.227} \\
            &   & MoDeGPT             & 0.089 & 0.000 & 0.208 & 0.099 \\
            \cmidrule(lr){2-7}
            & \multirow{6}{*}{70\%}
                & ASVD                & 0.078 & 0.006 & 0.119 & 0.068 \\
            &   & \textsc{SVD-LLM} v1 & 0.120 & 0.030 & 0.148 & 0.099 \\
            &   & \textsc{SVD-LLM} v2 & \textbf{0.129} & 0.033 & 0.129 & 0.097 \\
            &   & DoBi-SVD            & 0.115 & 0.000 & 0.116 & 0.077 \\
            &   & Basis Sharing       & 0.092 & \textbf{0.086} & \textbf{0.151} & \textbf{0.110} \\
            &   & MoDeGPT             & 0.113 & 0.000 & 0.137 & 0.083 \\
            \cmidrule(lr){2-7}
            & \multirow{6}{*}{60\%}
                & ASVD                & 0.102 & 0.002 & \textbf{0.115} & 0.073 \\
            &   & \textsc{SVD-LLM} v1 & 0.137 & \textbf{0.007} & 0.112 & 0.085 \\
            &   & \textsc{SVD-LLM} v2 & 0.098 & 0.004 & 0.112 & 0.071 \\
            &   & DoBi-SVD            & 0.113 & 0.000 & 0.110 & 0.074 \\
            &   & Basis Sharing       & 0.091 & 0.002 & 0.113 & 0.069 \\
            &   & MoDeGPT             & \textbf{0.146} & 0.000 & \textbf{0.115} & \textbf{0.087} \\
            \cmidrule(lr){2-7}
            & \multirow{6}{*}{50\%}
                & ASVD                & 0.094 & 0.000 & 0.114 & 0.069 \\
            &   & \textsc{SVD-LLM} v1 & \textbf{0.120} & 0.002 & 0.106 & 0.076 \\
            &   & \textsc{SVD-LLM} v2 & \textbf{0.120} & \textbf{0.003} & 0.111 & \textbf{0.078} \\
            &   & DoBi-SVD            & 0.104 & 0.000 & 0.111 & 0.072 \\
            &   & Basis Sharing       & \textbf{0.120} & 0.000 & 0.112 & 0.077 \\
            &   & MoDeGPT             & 0.074 & 0.000 & \textbf{0.117} & 0.064 \\
            \cmidrule(lr){2-7}
            & \multirow{6}{*}{40\%}
                & ASVD                & 0.120 & 0.000 & \textbf{0.116} & 0.079 \\
            &   & \textsc{SVD-LLM} v1 & 0.107 & \textbf{0.005} & 0.109 & 0.074 \\
            &   & \textsc{SVD-LLM} v2 & \textbf{0.142} & 0.000 & 0.114 & 0.085 \\
            &   & DoBi-SVD            & \textbf{0.142} & 0.000 & 0.115 & \textbf{0.086} \\
            &   & Basis Sharing       & 0.131 & 0.000 & 0.109 & 0.080 \\
            &   & MoDeGPT             & 0.089 & 0.000 & 0.115 & 0.068 \\
        \bottomrule
    \end{tabular}
\end{table*}

\subsection{BoolQ Label-Bias Audit}
\label{app:boolq_audit}

We further audit the anomalous BoolQ score of the MoDeGPT-compressed
Llama-3.1-8B checkpoint at an 80\% keep ratio. Two complete evaluation
reruns and an independent offline recomputation reproduce the same
accuracy of $0.4116$, ruling out evaluation nondeterminism. However,
although $62.2\%$ of the BoolQ ground-truth labels are ``yes,'' the
compressed checkpoint predicts ``yes'' for only $3.58\%$ of the
examples. Its class-conditional accuracies are $99.68\%$ on ``no''
examples and $5.56\%$ on ``yes'' examples, indicating a strong
output-label bias.

To distinguish checkpoint-level reproducibility from robustness to
calibration sampling, we regenerate compressed checkpoints using
different WikiText-2 calibration draws while holding the model,
compression ratio, and evaluation procedure fixed. As summarized in
Table~\ref{tab:boolq_calibration_sensitivity}, BoolQ varies substantially
across calibration draws, whereas the average over the other six
multiple-choice tasks remains nearly unchanged. The BoolQ range remains
$0.2064$ at 512 calibration sequences, showing that increasing the
calibration-set size alone does not eliminate this sensitivity. We
therefore interpret the $0.4116$ result as a reproducible,
calibration-dependent, task-specific label-bias failure rather than
general instability across the evaluation suite.

\begin{table}[t]
    \centering
    \caption{Calibration-draw sensitivity for MoDeGPT on Llama-3.1-8B
    at an 80\% keep ratio. Each range is the maximum minus the minimum
    score across three WikiText-2 sampling keys. ``Other-6'' denotes the
    average over ARC-Easy, ARC-Challenge, WinoGrande, PIQA, HellaSwag,
    and OpenBookQA.}
    \label{tab:boolq_calibration_sensitivity}
    \small
    \setlength{\tabcolsep}{8pt}
    \begin{tabular}{lcc}
        \toprule
        \textbf{Calibration sequences}
        & \textbf{BoolQ range}
        & \textbf{Other-6 range} \\
        \midrule
        128 & 0.2691 & 0.0030 \\
        256 & 0.1505 & 0.0020 \\
        \bottomrule
    \end{tabular}
\end{table}

\section{Platform Architecture and Artifact Management}
\label{app:platform_impl}



\subsection{Repository Organization}

\begin{figure}[ht]
    \centering
    \includegraphics[width=0.99\linewidth]{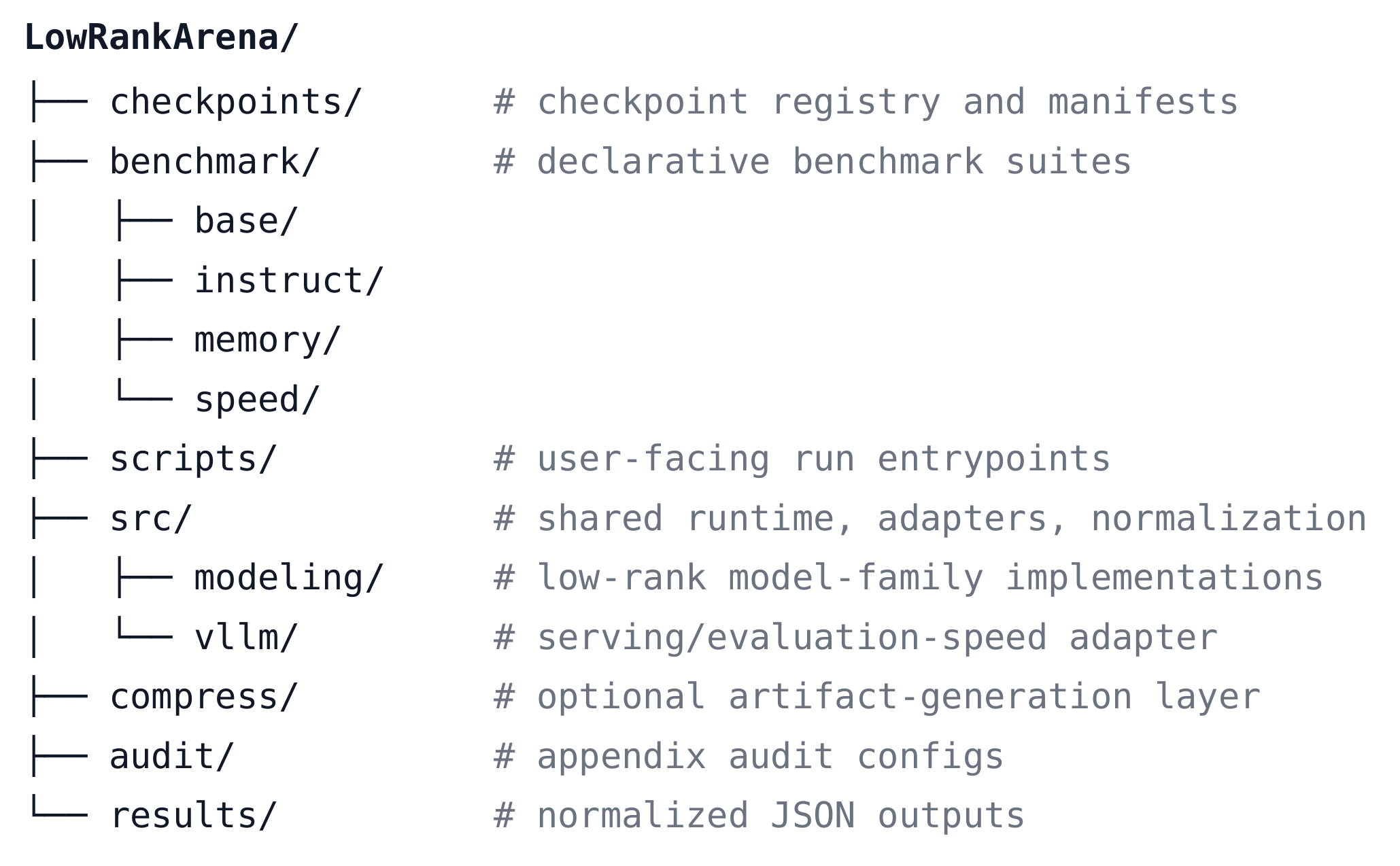}
    \caption{\textbf{LowRankArena repository organization.} Benchmark policy is specified declaratively in benchmark/, checkpoint metadata is centralized in checkpoints/, reusable execution logic lives in src/, and all runners emit normalized JSON results for downstream analysis.}
    \label{fig:repo_struct}
\end{figure}

\paragraph{Repository organization.}
Figure~\ref{fig:repo_struct} summarizes the released \benchname code layout.
The main design choice is to treat repository organization as part of the benchmark protocol rather than as incidental engineering: checkpoint records define the artifacts to be evaluated, YAML files define task and system workloads, shared runners implement loading and measurement, and outputs are stored in a normalized JSON schema. As a result, adding a future low-rank or compression method only requires exporting a compatible artifact and registering it through the same interface, while reusing the same task definitions, budget axis, precision policy, and serving stack.

\subsection{Adapting \benchname{} to New Methods}

\paragraph{Programmatic extension surface.}
\benchname{} exposes an \texttt{Arena} interface as the programmatic entrypoint for extending the benchmark. \texttt{Arena} wraps the checkpoint registry, sidecar manifests, loaders, evaluation runners, memory runner, speed runner, and reporting utilities behind a small set of operations. A method author can add a new artifact through \texttt{Arena.register} or \texttt{Arena.register\_manifest}, either as a local overlay for development or as a persisted registry row and manifest for release. Once registered, the artifact can be passed to the same \texttt{Arena.evaluate}, \texttt{Arena.memory}, and \texttt{Arena.speed} calls used by the released baselines.

This design makes extension artifact-centered: method-specific compression code may remain separate, but the benchmark-facing object is a registered checkpoint with explicit metadata. New methods therefore reuse the same task suites, budget axis, precision policy, loading path, serving adapters, and normalized JSON result schema rather than introducing bespoke evaluation scripts.

\subsection{Artifact Documentation and Traceability}
\label{app:artifact_documentation}

The released artifacts are documented through the
\href{https://huggingface.co/Duke-CEI-SVD/LowRankArena}
{LowRankArena Hugging Face repository}. Its checkpoint-to-result map
links each released checkpoint to its method, source model, keep ratio,
precision and calibration metadata, and normalized evaluation records.
The accompanying Croissant metadata and responsible-AI (RAI) datasheet
describe the artifact catalog, audit index, intended use, and known
limitations, while the license-inheritance matrix records the applicable
source-model, method-code, and redistributed-artifact terms. Upstream
license restrictions remain applicable and are not superseded by the
LowRankArena repository license.

The catalog also distinguishes \emph{main} from \emph{auxiliary}
artifacts. Main artifacts are the uniform-precision checkpoints and
results underlying Table~\ref{tab:leaderboard_tf} and the central
ranking and efficiency analyses. Auxiliary artifacts support calibration
sensitivity, instruction-tuned evaluation, mixed-precision or remapping
variants, cross-device robustness, and large-model feasibility; they are
released for traceability but do not enter the primary leaderboard or
determine its rankings.

\section{Discussion \& Limitations}
\label{app:limits}

Our contributions are twofold. First, we introduce LowRankArena, a standardized evaluation platform and a checkpoint zoo containing over 3~TiB of released artifacts; Figure~\ref{fig:placeholder} summarizes its design and reproducibility workflow. Second, we conduct a standardized empirical audit of five representative SVD methods through matched re-evaluation, rather than aggregating published results as a survey. Our primary audience is developers of SVD-based compression methods, with pruning, quantization, and ML-systems researchers as secondary users of the platform and artifacts. While \textsc{LowRankArena} provides a much-needed standardized audit for SVD-based LLM compression, our platform and analysis have several inherent limitations that highlight important avenues for future work. 

\paragraph{Methodological Scope and Precision.} 
To strictly isolate the capability and efficiency effects of low-rank factorization, our primary evaluation currently focuses on uniform-precision, post-training SVD compression. We do not evaluate hybrid schemes that tightly couple SVD with low-bit quantization (e.g., INT4/INT8) or unconstrained sparse masking. While these orthogonal techniques can push the capacity-efficiency Pareto frontier further, their inclusion would introduce confounding variables—such as precision casting overhead, quantization error, and specialized hardware support—that obscure fundamental rank-allocation dynamics. Future iterations of the arena aim to modularize the benchmark to accommodate multi-dimensional hybrid compression pipelines.

\paragraph{Hardware and Software Dependence.} 
Our end-to-end serving measurements capture the empirical reality of current systems, which means they are inevitably tied to specific hardware targets (A100, L40S, RTX A5000, H200) and the engineering maturity of the serving backend (vLLM). As demonstrated in our 70B feasibility tests (Table~\ref{tab:a1_large_model_feasibility}), current large-scale SVD compression is predominantly bottlenecked by software readiness—evidenced by CUDA eigensolver crashes and dependency failures—rather than theoretical resource limits. Consequently, as custom kernel support for low-rank operators matures in the broader system community, the practical speedups and feasibility boundaries reported in our current leaderboards will likely shift. 

\paragraph{Broader Impacts.} 
The proliferation of LLMs has led to unsustainable computational and environmental costs. By standardizing the evaluation of compression techniques and open-sourcing a 3+ TiB checkpoint zoo, \textsc{LowRankArena} mitigates the severe redundancy of researchers repeatedly reproducing fragmented baselines. It provides a stable, accessible ground truth that democratizes efficient LLM serving research. However, we caution that all model compression techniques, including SVD-based truncation, fundamentally alter the original representation space of the checkpoint. This perturbation may unpredictably impact downstream safety guardrails, alignment protocols, or inherent biases. We strongly encourage practitioners deploying compressed models from our repository to independently re-audit safety-critical capabilities before real-world deployment.



\end{document}